\documentclass[letterpaper, 10 pt, conference]{ieeeconf}  

\IEEEoverridecommandlockouts                              

\title{\LARGE \bf
GaussianVLM: Scene-centric 3D Vision-Language Models using Language-aligned Gaussian Splats for Embodied Reasoning and Beyond
}

\author{Anna-Maria Halacheva$^{1}$, Jan-Nico Zaech$^1$, Xi Wang$^{1,2,3}$, Danda Pani Paudel$^1$, Luc Van Gool$^{1}$\\
\\
$^1$INSAIT, Sofia University “St. Kliment Ohridski”,
$^2$ETH Zurich, $^3$TU Munich
}
\newcommand{\ourmodel}{GaussianVLM}

\usepackage[utf8]{inputenc} 
\usepackage[T1]{fontenc}    
\usepackage{hyperref}       
\usepackage{url}            
\usepackage{booktabs}       
\usepackage{amsfonts}       
\usepackage{nicefrac}       
\usepackage{microtype}      
\usepackage{graphicx}
\usepackage{subcaption}
\usepackage{amsmath}
\usepackage{colortbl}
\usepackage[table]{xcolor} 
\usepackage{makecell}

\begin{document}
\twocolumn[{%
\renewcommand\twocolumn[1][]{#1}%
\maketitle
\begin{center}
    \centering
    \includegraphics[width=\linewidth, trim={0cm 0cm 0cm 0cm}, clip]{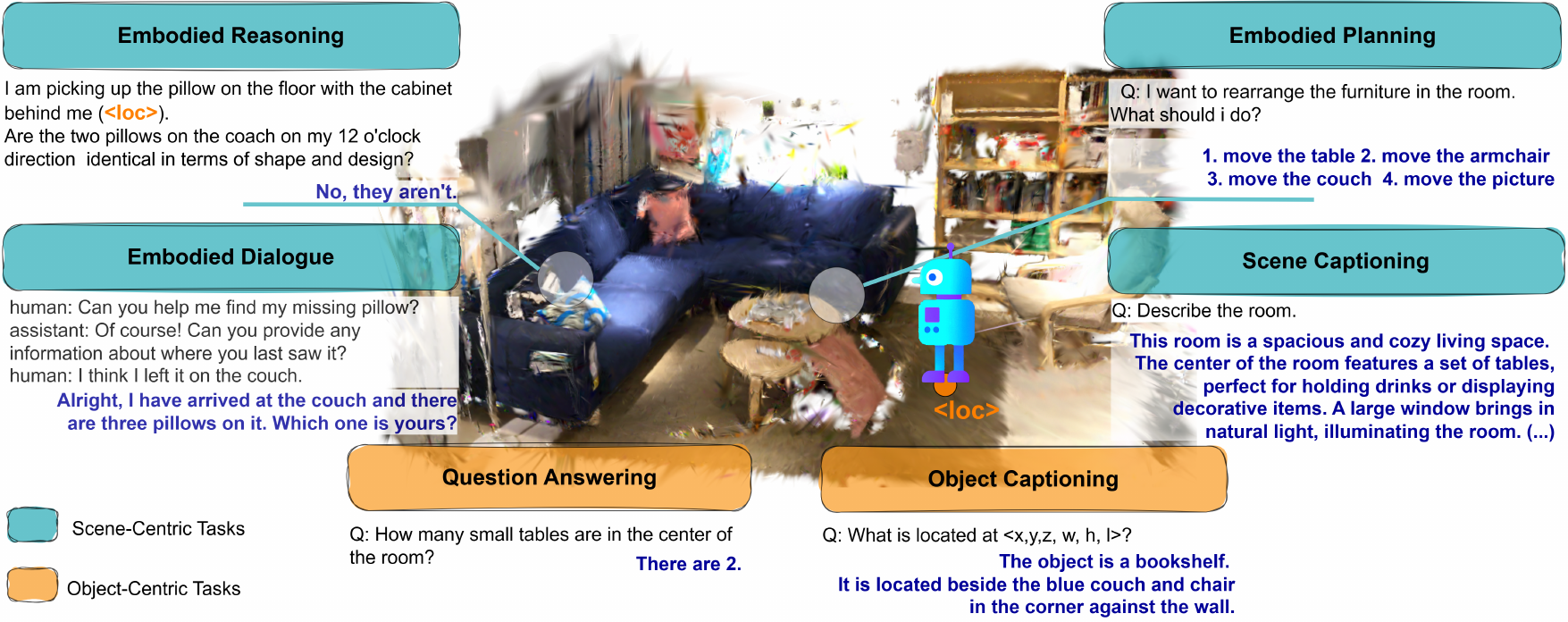}
    \captionof{figure}{The proposed \ourmodel~ performs comprehensive scene understanding in natural language for 3D scenes represented as Gaussian Splats. It adopts a fully scene-centric approach, building a global, language-augmented scene representation. This enables effective handling of both scene- and object-level tasks -- requiring multi-object reasoning, spatial understanding, global context, and fine-grained analysis -- suitable for embodied reasoning and beyond.    }
    \label{fig:teaser}
\end{center}
}]

\thispagestyle{empty}
\pagestyle{empty}


\begin{abstract}
    As multimodal language models advance, their application to 3D scene understanding is a fast-growing frontier, driving the development of 3D Vision-Language Models (VLMs). Current methods show strong dependence on object detectors, introducing processing bottlenecks and limitations in taxonomic flexibility.  To address these limitations, we propose a scene-centric 3D VLM for 3D Gaussian splat scenes that employs language- and task-aware scene representations. Our approach directly embeds rich linguistic features into the 3D scene representation by associating language with each Gaussian primitive, achieving early modality alignment. To process the resulting dense representations, we introduce a dual sparsifier that distills them into compact, task-relevant tokens via task-guided and location-guided pathways, producing sparse, task-aware global and local scene tokens. Notably, we present the first Gaussian splatting-based VLM, leveraging photorealistic 3D representations derived from standard RGB images, demonstrating strong generalization: it improves performance of prior 3D VLM (LL3DA \cite{chen2024ll3da}) five folds, in out-of-the-domain settings. 
    We provide \href{https://insait-institute.github.io/gaussianvlm.github.io/}{\textcolor{blue}{open access}} to all assets.
\end{abstract}

\section{Introduction}
To act intelligently in the physical world, embodied agents benefit from a rich, structured understanding of 3D scenes -- capturing not only objects but also spatial context, relationships, and semantics \cite{shorinwa2024splat,szot2021habitat,tidybot,rana2023sayplan}. Such scene understanding enables agents to move toward advanced tasks like embodied reasoning and planning, spanning multiple modalities \cite{lv2024robomp2,ma2022sqa3d,man2024situation,3dllm,Wijmans_2019_CVPR} .
While recent 3D VLMs have advanced towards addressing 3D vision-language tasks for embodied agents, they are predominantly object-centric, introducing a critical dependency on object detectors \cite{huang2023embodied,chen2024ll3da,huang2024chat,zhu20233d,3dllm}. This creates a mismatch with the core objective of generic scene understanding, forcing models into predefined granularities, limited taxonomies, and neglecting global context and spatial relationships \cite{Peng2023OpenScene,delitzas2024scenefun3d}. In this work, we propose to shift from object-centric to scene-centric representations by embedding language features directly into the spatial structure of the environment. Each element of the 3D scene, represented either as a point or a Gaussian splat, is enriched with continuous language features, e.g. CLIP \cite{radford2021learning}, SigLIP \cite{zhai2023sigmoid}. This allows us to construct a language-aligned scene representation without relying on predefined object categories. Our scene-centric 3D VLM thus can answer complex questions related to both objects and scenes, as shown in Fig.~\ref{fig:teaser}.

\begin{figure*}
    \centering
    \vspace{5pt}
    \includegraphics[width=1\linewidth]{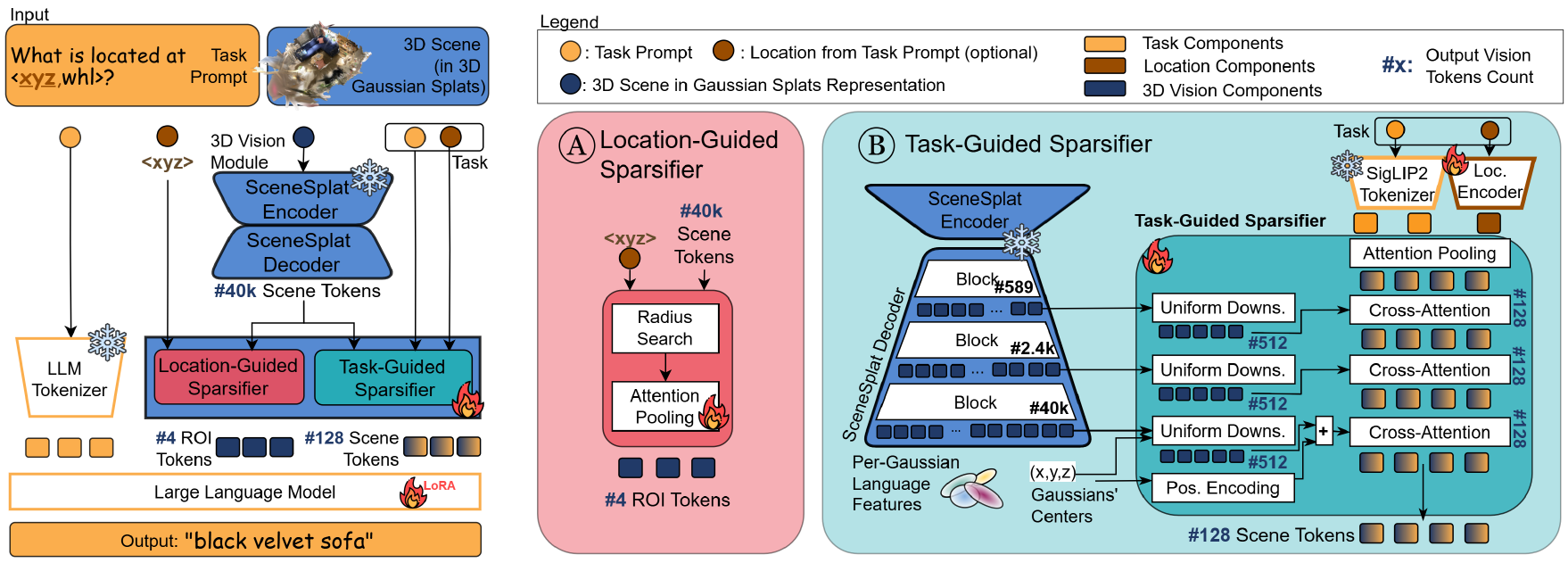}
    \caption{The \textbf{\ourmodel\ architecture} processes a user task prompt (query and optional location) and a 3D scene (Gaussian Splat representation).  A 3D vision module (SceneSplat Transformer) predicts per-Gaussian language features. These dense features are then sparsified by a dual sparsifier module.  The decoder's hidden states also inform the task-guided sparsifier.
    The dual sparsifier comprises: 1) a location-guided pathway that selects language features from Gaussians within a ROI around the task location, producing ROI tokens; and 2) a task-guided pathway that attends to dense scene tokens and SceneSplat decoder hidden states using task tokens (via cross-attention) to produce 128 task-selected scene tokens.  The resulting sparse scene representation (ROI tokens + task-selected tokens), along with the task tokens, is input to an LLM for response generation.}
    \label{fig:architecture}
    \vspace{-10pt}
\end{figure*}

However, directly embedding language features at the fine-granularity of the scene elements results in extremely dense representations in the tens of thousands tokens per scene. We argue that using the existing solutions, meaningfully understanding such representations via LLMs is a very challenging task -- due to the high density of high-dimensional language features. To address this, we introduce a dual sparsifier module that efficiently utlizes dense language representations while preserving semantic fidelity. The dual nature of our sparsifier has two pathways: task-guided and location-guided. The task-guided sparsifier selects scene tokens based on global task relevance, and the location-guided sparsifier retrieves fine-grained features conditioned on spatial cues in the task, as shown in Fig.~\ref{fig:architecture}. 
The location-based sparsifier selects the language features of the Gaussians within the Region-of-Interest (ROI) around the location from the task, reducing them to a few ROI tokens. The task-guided sparsifier takes as input the dense scene tokens and the task tokens, using the latter in cross-attention to guide the sparsification process. As a result, the dense features are reduced to 128 task-selected scene tokens. The obtained sparse scene representation, consisting of the ROI tokens and task-selected tokens, is passed together with the task tokens to an LLM for response generation.

Finally, we develop the first 3D VLM operating on Gaussian Splatting (GS) that naturally fuses geometry and appearance information \cite{li2025scenesplat}. Note that unlike point clouds, Gaussian splats capture detailed 3D textures -- in addition to the geometry -- which is necessary for generic 3D scene understanding of our interest. For the more, with the recent developments the high-quality 3DGS can be realistically acquired using only RGB cameras. We demonstrate that our model, \ourmodel, maintains strong task performance in real-world settings. We evaluate \ourmodel~ and a state-of-the-art (SOTA) point-cloud based VLM \cite{chen2024ll3da} on an in-house question-answering task for counting objects in ScanNet++ scenes \cite{yeshwanth2023scannet++}. On the utilized out-of-domain ScanNet++ scene representations, derived from RGB images, the GS-based \ourmodel~ outperforms the SOTA point cloud-based 3D VLM five folds in terms of accuracy (Tab.~\ref{tab:gs_vs_pc}).

We evaluate \ourmodel~ on a comprehensive suite of 3D vision-language tasks spanning both scene-centric (Tab.~\ref{tab:scene_centric}) and object-centric settings (Tab.~\ref{tab:object_centric}). Across the board, \ourmodel~ achieves state-of-the-art performance, outperforming the SOTA baselines \cite{chen2024ll3da,huang2023embodied} on every benchmark. Showing the advantages of scene-centrism, \ourmodel~ significantly outperforms previous methods on embodied scene-centric tasks, e.g., embodied reasoning (SQA3D \cite{ma2022sqa3d} 49.4\% vs. 47.0\% top-1 exact match) and substantially improving dialogue and planning metrics (e.g., +155.3 CIDEr in Embodied Planning \cite{3dllm}). Importantly, the detector-free \ourmodel~ also excels on object-centric benchmarks, e.g., achieving improved object captioning on Nr3D \cite{achlioptas2020referit_3d} (+15.0 METEOR, +9.3 ROUGE).

Overall, this work makes the following contributions: 
\begin{itemize}
    \item We introduce a fully scene-centric 3D VLM that achieves SOTA results, without requiring any dependencies on object detectors, on benchmark datasets for reasoning tasks required for embodied vision and beyond.
    \item We propose a dual sparsification mechanism to efficiently distill dense language-augmented scenes into compact, task-relevant representations, suitable for LLMs.
    \item We present the first language-grounded 3D VLM  directly operating on 3D Gaussian Splat representations.
\end{itemize}
\section{Related Work}
\subsection{Scene-Level Reasoning of Embodied Agents}
Early benchmarks in embodied question answering (EQA) \cite{Das_2018_CVPR, Wijmans_2019_CVPR} pioneered tasks requiring agents to reason from egocentric observations, primarily focusing on situated, navigation-oriented challenges. Subsequent research expanded this scope to include multi-hop and commonsense reasoning \cite{ma2022sqa3d}, as well as embodied planning and dialogue tasks \cite{3dllm}. Early solutions adapted architectures like MCAN \cite{yu2019deep} and ClipBERT \cite{lei2021less}, with ScanQA \cite{scanqa_22_cvpr} introducing 3D scene-grounded QA via explicit reconstructions. This progression has culminated in generalist 3D VLMs \cite{huang2023embodied,zhu2025unifying,chen2024ll3da,zhu20233d} unifying 3D scene understanding, reasoning, and planning.

\subsection{3D Scene Tokenization}
For effective VLMs, 3D scene tokenization transforms complex geometry into language-processable, semantically rich representations. Two prevalent strategies exist:

\textbf{Object-Level Tokenization.} 
A common paradigm \cite{huang2023embodied,huang2024chat,li2025m3dbench,chen2024grounded3dllm,zhang2024spartun3d,zhu20233d,wang2023chat3d} involves detecting individual objects, extracting their point clouds, and independently encoding them with a 3D encoder to generate object-level tokens. This method, while semantically intuitive, is limited by object detector performance and neglects vital scene context (e.g., room layout, walls).

\textbf{Region-Based Tokenization.} 
Another approach \cite{zhi2024lscenellm,zhu2025unifying} encodes the entire scene into per-point features, then groups these points into a fixed number of regions (e.g., via kNN \cite{zhi2024lscenellm} or graph-based segmentation \cite{zhu2025unifying}). Averaging features within these regions creates region-level tokens, capturing broader context at reduced granularity. However, this risks over-smoothing by collapsing diverse information into single tokens. Additionally, predefining the number of regions is challenging: too many can introduce irrelevant data and increase cost, while too few may lose fine-grained details.

In contrast, we introduce \textbf{language-guided scene tokenization}, an approach that dynamically re-tokenizes the scene based on linguistic input and per-point/per-Gaussian language features. By leveraging language to direct the tokenization, our method ensures that the resulting tokens focus on the scene regions most pertinent to the current task.






\subsection{Vision-Language-Aligned 3D Scene Understanding}
Integrating language into 3D scene understanding introduces challenges, particularly in (1) achieving effective cross-modal alignment \cite{huang2023embodied,chen2024ll3da}, and  (2) ensuring semantically rich vision features \cite{zhu2025unifying}.

\textbf{Text-Vision Alignment.}
Prior work commonly aligns 3D visual features with language by projecting each modality independently into a shared embedding space \cite{huang2023embodied, huang2024chat,fu2025scene,wang2023chat3d}. However, this often results in weak alignment due to the largely separate processing of the two modalities. 
In line with 2D vision-language models \cite{li2023blip}, other approaches employ learnable query tokens that attend to both visual and textual features, separately, \cite{zhu2025unifying, chen2024ll3da}, aiming for information fusion. Nevertheless, these query-based methods frequently refine visual features before language interaction, limiting the language's impact on the initial visual encoding. Critically, a shared limitation across these strategies is that the 3D encoder features are generated without incorporating any language or task-relevant semantic cues, ultimately leading to a shallow alignment \cite{tschannen2025siglip}.  Our approach, in contrast, ensures strong text-vision alignment by embedding language features directly into the fine-grained spatial structure of the 3D scene.

\textbf{Vision Feature Quality.}
Recent efforts in 3D scene representation and sparsification have aimed to improve VLM performance by increasing the vision feature expressiveness.
Many approaches leverage multi-modal visual data (2D images, point clouds, meshes) \cite{huang2024chat,zhu2025unifying,3dllm} for rich scene information, yet they are computationally intensive and architecturally complex, often also with inefficient, task-agnostic sparsification. Region highlighting techniques \cite{zhi2024lscenellm,mei2025PerLA,chen2024ll3da} attempt to emphasize key regions alongside a global scene representation, but the persistent use of dense global representations limits scalability and and attentional focus. We avoid these limitations by (a) using easy-to-obtain expressive language-aligned features \cite{tschannen2025siglip} as our scene representation, and (b) generating all scene tokens conditioned on the task.

\section{Method}
We introduce \ourmodel, a 3D VLM for indoor scene understanding. Given a 3D scene represented as Gaussian splats and a natural language prompt, \ourmodel~ fuses language and 3D vision at multiple stages to generate a textual response. Notably, \ourmodel~ is the first to leverage Gaussian splats as the 3D scene representation, and function exclusively in the language space, achieving this object detector-free. \ourmodel~relies on three key innovations: (1) a language-aware Gaussian splatting backbone \cite{li2025scenesplat} that predicts language features for each Gaussian, enabling direct language-based alignment between the scene and the prompt; (2) a task-guided sparsifier module generating a sparse scene representation by performing task-aware re-tokenization of the dense 3D backbone output; and (3) a location-guided sparsifier module for detector-free extraction of Region-of-Interest (ROI) information. We detail the \ourmodel~ and the sparsifier components in the subsequent sections.

\subsection{\ourmodel}
Unlike previous approaches that rely on purely visual representations, our method integrates a 3D transformer that produces inherently language-grounded vision features. Specifically, we adopt SceneSplat \cite{li2025scenesplat} as our 3D vision module. SceneSplat processes scenes represented via Gaussian splats and predicts a SigLIP2 \cite{tschannen2025siglip} language feature for each Gaussian end-to-end. To sparsify the resulting dense language features with a task-awareness, we introduce a dual sparsifier  module. The sparsifier takes as input the dense language features and outputs sparse task-aware tokens. The sparse scene tokens are projected from the SigLIP2 space into the LLM space via a single linear projection. The resulting vision tokens are then concatenated with the user task tokens, tokenized via the LLM's tokenizer, and input into a frozen LLM augmented with Low-Rank Adaptation (LoRA) \cite{hu2022lora}. The LLM autoregressively generates responses to the user query, conditioned jointly on both visual and textual context. \ourmodel~ (OPT-1.3B \cite{zhang2022opt} as LLM) has a size of 
1.8B parameters out of which 19M are learnable. 

\textbf{Training Objective. }Similarly to many VLM training protocols \cite{vis_inst_tuning,huang2023embodied,zhang2024spartun3d}, we follow a two-stage training with alignment and fine-tuning phase. During the alignment phase we freeze the 3D backbone and LLM tokenizer, training the sparsifier modules and the transformer for textual alignment of the vision tokens. The LLM  is adapted using LoRA. Both stages share a unified training objective. Following \cite{huang2023embodied,NEURIPS2020_1457c0d6,raffel2020exploring}, we use a prefix
language modeling, where the model is conditioned on an input prefix and trained to autoregressively generate the target continuation:
\begin{equation}
    \mathcal{L}(\theta, \mathcal{B}) = - \sum_{\{s_{\text{prefix}},s_{\text{gt}}\}\in B}\sum_{t=1}^{|s_{\text{gt}}|} \log p_\theta \left( s_{\text{gt}}^{(t)} \mid s_{\text{gt}}^{(<t)}, s_{\text{prefix}} \right),
\end{equation}
with $\theta$ as the model parameters, $\mathcal{B}$ - a batch of samples of prefix input $s_{\text{prefix}}$ (task prompt and vision tokens), and ground truth response $s_{\text{gt}}$. $s_{\text{gt}}^{(t)}$ denotes the $t$-th token in the ground truth response sequence. 

To enhance spatial grounding, we initially pre-train the task-guided sparsifier on understanding 3D location features. We leverage an object captioning task in which the model is provided the 3D coordinates of a labeled object instance and trained to generate a visual token embedding similar to the embeddings of the corresponding label token. The location is encoded through learnable Fourier embeddings (Eq. \ref{eq:fourier}) to a single feature and passed to the sparsifier. The label text is embedded using the SigLIP-2 tokenizer. This pre-training stage uses a one-sided contrastive objective \cite{radford2021learning}, encouraging the output embedding of the task-guided sparsifier $s_i$ to match its corresponding label embedding $l_i$, while being distant from all other labels $l_j$ ($j \neq i$):

\begin{equation}
\mathcal{L}{\text{contrast}} = -\log \frac{\exp(s_i^\top l_i / \tau)}{\sum{j=1}^{N} \exp(s_i^\top l_j / \tau)},
\end{equation}
where $s_i$ is the output \textbf{s}cene token of the sparsifier for the $i$-th instance, $l_i$ is the SigLIP-2 embedding of the corresponding \textbf{l}abel, $N$ is the number of labels in the batch, and $\tau$ is a temperature hyperparameter which we set to 0.07. 

\subsection{Dual Sparsifier}
\textbf{Task-Guided Sparsification.} SceneSplat processes 3D Gaussians into a dense sequence of tokens (one per Gaussian). Following established practices \cite{chen2024ll3da, zhi2024lscenellm, mei2025PerLA, chen2023end}, sampling 40k Gaussians yields a corresponding 40k output tokens, originating from different SceneSplat decoder layers (specifically, 589, 2.4k, and 40k). To address the computational demands of this dense representation and prioritize task-relevant information, we introduce a novel task-guided sparsification module. This module re-tokenizes the dense scene representation into a more compact, sparse one by selectively attending to the most important visual features based on the textual query.

Our sparsifier employs the language task to generate queries that guide the filtering of visual input via depth-wise cross-attention \cite{pizero}. This task-guided sparsification is applied iteratively to the output of each SceneSplat decoder layer, enabling a dynamic and context-aware reduction of visual information.

To mitigate the computational overhead of cross-attention on a large number of tokens, we first apply a simple uniform downsampling strategy to reduce the representation to 512 tokens per decoder layer. Our ablation study (Sec.~\ref{sec:ablation}) demonstrates the sufficiency of this efficient approach, negating the need for more complex initial downsampling methods like kNN used in other models \cite{mei2025PerLA,zhi2024lscenellm}. Subsequently, we further sparsify to 128 tokens by performing cross-attention between the tokens of the user's prompt (tokenized using SigLIP2, consistent with SceneSplat's language features) and these weakly-sparsified 3D features. For any spatial locations \( \text{loc}_{xyz} \in \mathbb{R}^3 \) mentioned in the prompt, we encode them using learnable Fourier embeddings \cite{chen2024ll3da}:
\begin{equation}
   \text{pos}(\text{loc}_{xyz}) = [\sin (2\pi \, \text{loc}_{xyz} \cdot B) \; ; \; \cos (2\pi \, \text{loc}_{xyz} \cdot B)] 
   \label{eq:fourier}
\end{equation}
where \( B \in \mathbb{R}^{3 \times (d/2)} \) is a learnable matrix. If the prompt includes a bounding box, we extract \( \text{loc}_{xyz} \) as its center. 

To generate queries for cross-attention, we apply attention pooling to the embeddings of the task tokens, resulting in a fixed set of 128 query vectors. 
Corresponding to each SceneSplat decoder block is a cross-attention sparsifier block. The initial layer of these blocks performs cross-attention between the SceneSplat visual tokens and the task tokens. The resulting intermediate visual features are then processed through subsequent layers, further sparsifying the scene features and refining their semantic alignment with the language in a depth-wise manner. This process yields language-aware vision tokens integrating global scene understanding from the earlier decoder layers with instance-level awareness derived from the per-Gaussian language features.

In the final sparsifier layer, we additionally inject positional information by encoding the center of the downsampled 512 Gaussian splats using Eq.~\ref{eq:fourier}. This step is crucial for instilling position awareness into the otherwise location-agnostic Gaussian language features.

\textbf{Location-Guided Sparsification.} For tasks that provide a location, such as object captioning, we introduce an ROI magnifier. This module extracts features from a spherical region around the object location indicated by the prompt. For click locations, we use the click's xyz point; for bounding boxes, we use the center. Given a location point, we select neighboring points within a 15cm radius, chosen to focus on small objects. If no points are captured in the ROI, we iteratively increase the radius by 15cm until the ROI is not empty. We then apply attention pooling to the language features of these selected points to generate 4 ROI tokens summarizing the region.




\section{Experiments}
\subsection{Dataset}
We evaluate our model under the LL3DA, a SOTA 3D VLM, training protocol \cite{chen2024ll3da}. We also evaluate on embodied reasoning (SQA3D \cite{ma2022sqa3d}), a popular 3D VLM benchmark \cite{huang2023embodied,zhu2025unifying}, where we follow the LEO \cite{huang2023embodied} training protocol.

\textbf{LL3DA Training Protocol.} For the LL3DA protocol, we follow their one-stage joint training procedure. Training is performed on ScanRefer \cite{chen2020scanrefer} (object captioning), ScanQA \cite{scanqa_22_cvpr} (general question-answering (QA)), Nr3D \cite{achlioptas2020referit_3d} (object captioning), and the ScanNet subset of 3D-LLM \cite{3dllm} (diverse scene-centric tasks), focusing on multitask learning.

\textbf{LEO Training Protocol.}
In the LEO setting, we adopt a two-phase training strategy with alignment and instruction tuning. To maintain compatibility with our scene-centric design and the LL3DA setup, we restrict training to the ScanNet subset of the LEO dataset.
We align the visual and language modalities using the ReferIt3D dataset \cite{achlioptas2020referit_3d} providing detailed object captions. This phase helps the model ground linguistic features directly into the 3D scene representation. During the second stage, the model is further trained to follow natural language instructions across multiple tasks using the SQA3D (situated QA) , ScanRefer, and ScanQA datasets.

\subsection{Tasks}
We evaluate our model on a diverse set of 3D vision-language tasks drawn from the LL3DA and LEO benchmarks. These tasks fall into two broad categories: object-centric and scene-centric, reflecting differing demands on spatial grounding and semantic abstraction.

\textbf{Object-Centric Tasks} require reasoning about discrete objects in the scene, often relying on explicit object annotations or localized queries. Those tasks include:

\textit{Object Captioning.} We use ScanRefer \cite{chen2020scanrefer}, and Nr3D \cite{achlioptas2020referit_3d} for evaluating object captioning. Each training instance provides a natural language expression referring to a specific object in the scene. Using the annotated instance IDs, we extract the corresponding 3D bounding box and use its center as the target location. The model is prompted to generate a caption for the object at this location, conditioned on the full scene representation. To encourage linguistic diversity, we use GPT-4o to generate 40 paraphrased variants per prompt with varied syntax and vocabulary.

\textit{Object-Centric Question Answering.}
We use ScanQA \cite{scanqa_22_cvpr}, which includes questions about object attributes, counts, and presence (e.g., “What is the color of the chair?”). As these questions typically target individual entities rather than spatial relationships or global context, the dataset aligns with object-centric evaluation.

\textbf{Scene-Centric Tasks}, in contrast, require holistic reasoning about the environment, its layout, and the agent’s situated context—without reducing the scene to individual object tokens.
The \textit{Situated Question Answering (SQA3D) \cite{ma2022sqa3d}} requires the model answering answer spatial or functional questions grounded in the scene (e.g., “What is on my left?”), given a situational context (e.g., “I am washing my hands”). These questions require understanding the scene’s layout, affordances, and agent-relative positioning.
For the \textit{Embodied Planning \cite{3dllm}} task, the model generates high-level plans to complete tasks, leveraging the full scene structure to identify relevant objects and transitions. 
In \textit{Scene Captioning \cite{3dllm}}, the model produces free-form descriptions summarizing the entire scene, requiring it to integrate geometry, object presence, and semantics into coherent language.
\textit{Embodied Dialogue \cite{3dllm}} introduces an interactive setting, where the model answers context-aware questions or participates in a dialogue about the scene, requiring dynamic grounding and multi-turn understanding.

\subsection{Metrics}
For scene-centric tasks, where captions and answers typically encompass diverse and richly descriptive content, we report standard metrics including CIDEr \cite{vedantam2015cider}, BLEU-4 \cite{papineni2002bleu}, METEOR \cite{banerjee2005meteor}, ROUGE \cite{chin2004rouge}, exact-match accuracy, and Sentence-BERT \cite{reimers-gurevych-2019-sentence} similarity. 

For object-centric tasks, we exclude BLEU-4 and CIDEr. BLEU, a precision-based metric, and CIDEr are overly sensitive to superficial n-gram overlap, rendering them unsuitable for evaluating long-form object captions \cite{anderson2016spice}. These captions extend beyond simple object naming (e.g., "a bed") to include context (e.g., "the bed is rectangular and has a white bedspread. it is located between the end table with the lamp on it (...)"). Consequently, BLEU and CIDEr can assign misleadingly high scores to captions that correctly describe the scene context but identify the wrong object. This occurs because these metrics reward overlapping phrases and frequent n-grams, even with an incorrect core referent. In contrast, we employ METEOR, ROUGE, and Sentence-BERT similarity, which offer superior handling of semantic alignment and partial matches \cite{anderson2016spice,reimers-gurevych-2019-sentence}. Specifically, METEOR incorporates synonym matching and alignment at the word and phrase level. ROUGE captures structural similarity and emphasizes recall without over-rewarding redundant context. Sentence-BERT directly evaluates semantic similarity in embedding space for robustness to paraphrasing.

\subsection{Implementation Details}
\label{sec:implement}
Following prior work, we represent each 3D scene using 40k randomly sampled Gaussians from the GaussianWorld \cite{li2025scenesplat} Gaussian splats scene. For the language model, we adopt OPT-1.3B \cite{zhang2022opt} for LL3DA settings and Vicuna-7B \cite{chiang2023vicuna} for LEO, as per their respective training protocols. Both LLMs are loaded in float16 for memory efficiency and fine-tuned using LoRa. Our training procedure adheres to the standard protocols: 5 epochs of alignment followed by 10 epochs of instruction tuning for LEO, and 32 epochs for LL3DA. Training completes in under one day on 8 A100-80 GPUs. Additionally, we pre-train our task-guided sparsifier on the object captioning task for 5 epochs
We employ the AdamW optimizer with a weight decay of 0.1 and a cosine annealing learning rate schedule, decaying from ($10^{-4}$) to ($10^{-6}$). Evaluation is performed every 8 epochs for LL3DA and every epoch for LEO.


\subsection{Results and Analysis}
\begin{table*}[ht]
\vspace{5pt}
\small
\centering
\resizebox{\textwidth}{!}{
\begin{tabular}{l|c|c|c|c|c|c|c|c|c|c|c|c|c|c|c}
\toprule
& \multicolumn{5}{c}{\textbf{Embodied Dialogue}} & \multicolumn{5}{c}{\textbf{Embodied Planning}} & \multicolumn{5}{c}{\textbf{Scene Captioning}} \\
\cmidrule(lr){2-6} \cmidrule(lr){7-11} \cmidrule(lr){12-16}
& Sim & C & B-4 & M & R & Sim & C & B-4 & M & R & Sim & C & B-4 & M & R \\
OPT-1.3B \cite{zhang2022opt} &-& 0.31 &0.23 & 5.62 &4.83 & - &0.16& 0.13& 0.24 &3.56 &- &0.0& 0.84 & 8.40 &11.7\\
OPT-2.7B  \cite{zhang2022opt} &-&0.38& 0.39 &7.38 &6.28 & -&0.10& 0.26 & 3.59& 4.35&-&  0.11&  0.00 &  6.60&  12.32\\
OPT-6.7B  \cite{zhang2022opt} &-&0.25 &0.43 & 6.88 &6.16& -&0.00& 0.28&  3.65& 3.94&- & 0.06 & 1.13& 8.99& 16.96\\
LLAMA-7B  \cite{touvron2023llama} & - & 0.27& 0.50 & 7.81 &6.68& -&0.04& 0.29 & 3.53 &4.71&- & 0.2 & 0.92 &7.00 &12.31
\\ 
\midrule
LL3DA* \cite{chen2024ll3da} & 48.2 & 145.9 & 22.2 & 40.9 & 36.7 & 50.2 & 65.1 & 7.1 & 20.8 & 32.2 & \textbf{66.4} & 0.2 & 3.0 & 19.4 & 18.4 \\
\textbf{\ourmodel~(Ours)} & \textbf{72.3} & \textbf{270.1} & \textbf{31.5} & \textbf{55.7} & \textbf{48.6} & \textbf{59.0} & \textbf{220.4} & \textbf{20.3} & \textbf{44.5} & \textbf{48.0} & 65.8 & \textbf{0.8} & \textbf{6.4} & \textbf{23.5} & \textbf{21.1} \\
\bottomrule
\end{tabular}}
\caption*{(a) LL3DA Scene-Centric Benchmarks. We compare 3D VLMs and frozen LLMs, following \cite{chen2024ll3da}. Our method, \ourmodel, outperforms all baselines by a large margin.}

\begin{minipage}{0.48\textwidth}
\vspace{10pt}
\centering
\resizebox{\textwidth}{!}{
\begin{tabular}{l|c|c|c|c|c}
\toprule
&\multicolumn{5}{c}{\textbf{SQA3D}} \\
\cmidrule(lr){2-6} 
\textbf{} & EM1 & C & B-4 & M & R \\
\midrule
GPT3 \cite{brown2020language} & 41.0 & - & - &- &-\\
ClipBERT \cite{lei2021less}& 43.3  & - & - &- &-\\
SQA3D \cite{ma2022sqa3d} & 46.6  & - & - &- &-\\
\midrule
3D-VisTA \cite{zhu20233d} & 48.5 & - & - &- &-\\
PQ3D \cite{zhu2025unifying} & 47.1& - & - &- &-\\
LEO* \cite{huang2023embodied} & 47.0 & 124.7 & 9.4 & 25.5 & 48.4 \\
\textbf{\ourmodel~(Ours)} & \textbf{49.4} & \textbf{129.6} & \textbf{17.1} & \textbf{26.4} & \textbf{50.2} \\
\bottomrule
\end{tabular}}
\caption*{(b) LEO Scene-Centric Benchmarks}
\end{minipage}
\hfill
\begin{minipage}{0.49\textwidth}
\vspace{-30pt}
\caption{Evaluation of SOTA 3D VLMs on\textbf{ scene-centric} 3D vision-language tasks.  (a) Results on the scene-centric benchmarks from LL3DA. (b) Results on the scene-centric benchmarks from LEO. We report results from specialist models (top) and generalist 3D VLMs (bottom). (*): reproduced.  Evaluation metrics include CIDEr (C), BLEU-4 (B-4), METEOR (M), ROUGE (R), Sentence Similarity (Sim), and Top-1 Exact Match (EM1).}
\label{tab:scene_centric}
\end{minipage}
\end{table*}

\begin{table}[ht]
\small
\vspace{-10pt}
\centering
\begin{tabular}{@{\hskip 0mm}l@{\hskip 0.7mm}|@{\hskip 0.7mm}c@{\hskip 0.7mm}|@{\hskip 0.7mm}c@{\hskip 0.7mm}|@{\hskip 0.7mm}c@{\hskip 0.7mm}|@{\hskip 0.7mm}c@{\hskip 0.7mm}|@{\hskip 0.7mm}c@{\hskip 0.7mm}|@{\hskip 0.7mm}c@{\hskip 0.7mm}|@{\hskip 0.7mm}c@{\hskip 0.7mm}|@{\hskip 0.7mm}c@{\hskip 0.7mm}|@{\hskip 0.7mm}c@{\hskip 0mm}}
\toprule
& \multicolumn{3}{c}{\textbf{ScanRefer}} & \multicolumn{3}{c}{\textbf{ScanQA}} & \multicolumn{3}{c}{\textbf{Nr3D}} \\
\cmidrule(lr){2-4} \cmidrule(lr){5-7} \cmidrule(lr){8-10} 
& Sim &  M & R & EM1 & M & R & Sim &  M & R \\
\midrule
Scan2Cap \cite{chen2021scan2cap}& - & 21.4 & 43.5&-&-&-&-&-&-\\
\makecell[l]{VoteNet+\\MCAN \cite{yu2019deep}}& - &-&- &  17.3 &11.4 &29.8 & - &- & -\\
ScanQA \cite{scanqa_22_cvpr} & - &-&-& - & 13.14& 33.3&-&-&-\\
\midrule
3D-LLM \cite{3dllm}& - &13.1 &33.2 &\textbf{19.3} &13.8 &34.0&-&-&- \\
3D-VLP \cite{yang20243d} & - &-&-& - & 13.5& 34.5&-&-&-\\
Scene-LLM \cite{fu2025scene} & - &21.8 &45.6 & - &  15.8 & -&-&-&- \\
LL3DA* \cite{chen2024ll3da} & 55.9 & 51.6 & 54.8 & 14.3 & 22.8 & 34.7 & 48.1 & 5.8 & 9.9 \\
\makecell[l]{\textbf{\ourmodel}\\\textbf{(Ours)}} &\textbf{59.1} &  \textbf{52.4} &  \textbf{57.4} &  14.4  & \textbf{22.9} &  \textbf{34.8}& \textbf{48.2} & \textbf{20.8} & \textbf{19.2}\\
\bottomrule
\end{tabular}
\vspace{5pt}
\caption{Evaluation on \textbf{object-centric} LL3DA benchmarks. We report both specialist models (top), and 3D VLMs (bottom). (*): reproduced.  Models focusing on grounding (3D-LLM, 3D-VLP, Scene-LLM) and specialist models were not reproduced due to differing objectives.}
\label{tab:object_centric}
\vspace{-10pt}
\end{table}
The evaluation results, shown in Tab. \ref{tab:scene_centric} and Tab. \ref{tab:object_centric}, highlight \ourmodel's effectiveness across both training protocols and task types. On the scene-centric SQA3D benchmark, \ourmodel~ achieves an exact match accuracy of 49.4\%, surpassing LEO’s 47.0\% by 2.4 percentage points. Under the LL3DA protocol, \ourmodel~ significantly improves embodied dialogue and planning tasks, with CIDEr scores increasing from 145.9 to 270.1 (+124.2) and from 65.1 to 220.4 (+155.3), respectively, demonstrating enhanced multi-object reasoning and spatial context understanding. In object-centric evaluations (Tab. \ref{tab:object_centric}), \ourmodel~ achieves comparable (ScanQA) or superior results (e.g., Nr3D with a METEOR score of 20.8 versus 5.8) to existing methods, despite not employing object detectors.

\subsection{Real-World Generalization}
To assess generalization to data obtainable in realistic real-world settings, we also evaluate \ourmodel~ and LL3DA on scene representations derived from RGB image data. Unlike traditional point cloud-based VLMs, which often rely on laser-scanned geometry, our model is trained on photorealistic Gaussian splats, potentially offering better robustness to less structured inputs. For this experiment, we use the ScanNet++ \cite{yeshwanth2023scannet++} (validation split), which is  out-of-domain (OOD) for our setup consisting exclusively of ScanNet scenes. Specifically, we utilize GaussianWorld's \cite{li2025scenesplat} ScanNet++ scenes, generated from RGB data, for our 3DGS representation, while the point cloud baseline (LL3DA) employs COLMAP \cite{schonberger2016pixelwise,schonberger2016structure} reconstructions from ScanNet++. To address the lack of suitable benchmarks on ScanNet++, we introduce a novel object counting question-answering dataset. This dataset, automatically constructed using ScanNet++ segmentation annotations, comprises 1000 question-answer pairs focused on object counts. We exclude non-object categories to ensure focused evaluation. We evaluate our model and LL3DA on this OOD dataset using standard question-answering evaluation protocols, specifically Exact Match, ROUGE, METEOR, CIDEr, as well as Accuracy.  The results reveal a significant performance advantage for our Gaussian splat-based model, outperforming the point cloud-based SOTA VLM (LL3DA) by 474\% in accuracy on the GS scenes (Tab.~\ref{tab:gs_vs_pc}). Further details on dataset construction and statistics are provided in the supplementary material.

\begin{table*}[h!]
\vspace{5pt}
\centering
\small
\begin{tabular}{l|c|c|c|c|c}
\toprule
\textbf{Model} & \textbf{Accuracy (\%)} & \textbf{EM} & \textbf{CIDEr} & \textbf{METEOR} & \textbf{ROUGE} \\
\midrule
LL3DA \cite{chen2024ll3da}        & 4.2    & 1.5   & 54.4     & 25.5 & 26.8 \\
\textbf{\ourmodel~(Ours)}   & \textbf{24.1} &\textbf{ 9.3} & \textbf{120.0 }     & \textbf{35.2} & \textbf{47.3} \\
\midrule
Improvement $\%$ & +474.0\%  & +520.0\%  & +120.6\% & +38.0\% & +76.5\% \\
\bottomrule
\end{tabular}
\vspace{-5pt}
\caption{Evaluation of QA on object counts on the out-of-domain ScanNet++ validation scenes.}
\label{tab:gs_vs_pc}
\end{table*}

\begin{table*}[ht]
\small
\centering

\begin{tabular}{@{\hskip 0mm}l@{\hskip 0.8mm}|@{\hskip 0.8mm}c@{\hskip 0.8mm}|@{\hskip 0.8mm}c@{\hskip 0.8mm}|@{\hskip 0.8mm}c@{\hskip 0.8mm}|@{\hskip 0.8mm}c@{\hskip 0.8mm}|@{\hskip 0.8mm}c@{\hskip 0.8mm}|@{\hskip 0.8mm}c@{\hskip 0.8mm}|@{\hskip 0.8mm}c@{\hskip 0.8mm}|@{\hskip 0.8mm}c@{\hskip 0.8mm}|@{\hskip 0.8mm}c@{\hskip 0.8mm}|@{\hskip 0.8mm}c@{\hskip 0.8mm}|@{\hskip 0.8mm}c@{\hskip 0.8mm}|@{\hskip 0.8mm}c@{\hskip 0.8mm}|@{\hskip 0.8mm}c@{\hskip 0.8mm}|@{\hskip 0.8mm}c@{\hskip 0.8mm}|@{\hskip 0.8mm}c@{\hskip 0mm}}
\toprule
\multicolumn{16}{c}{\textbf{Scene-Centric Tasks}} \\
\toprule
 & \multicolumn{5}{c}{\textbf{Embodied Dialogue}} & \multicolumn{5}{c}{\textbf{Embodied Planning}} & \multicolumn{5}{c}{\textbf{Scene Captioning}} \\
\cmidrule(lr){2-6} \cmidrule(lr){7-11} \cmidrule(lr){12-16}
& Sim & C & B-4 & M & R & Sim & C & B-4 & M & R & Sim & C & B-4 & M & R\\
\midrule
(1) No Vision Tokens & 13.5& 0& 0& 1.1& 0& 15.7& 0& 0& 0.4& 0& 0.3& 0& 0& 0.9& 0.4\\
(2) No Scene Tokens  & 69.3 & 234.9& 28.0& 52.0& 45.3& 54.1& 156.1& 3.9& 36.9& 40.1& 61.5& 0.7& 1.3& 15.4& 17.4\\
(3) No ROI Tokens  & 68.9& 233.4& 28.1& 52.0& 44.9& 56.8& 195.0& 12.0& 41.0& 44.8& 63.4& 2.5& 3.1& 19.6& 20.9\\
(4) Only Vision Tokens & 34.7 & 67.5 & 8.8 & 24.9& 19.9  & 37.0 & 46.6 & 4.1 & 21.1 & 25.8 & 37.8 & 0 & 0 & 0.2 & 0.3\\
\midrule
(5) No Depth-Wise CA & 71.2 &  269.1 & 30.9 & 55.2 & 48.3 & 58.3 & 209.3 &18.6 & 44.2 & 47.9 & 64.4 & \textbf{2.4} & 4.9 & 21.8 & \textbf{21.1}\\
(6) No Text-Guidance & 71.4 & 267.0 & 31.3 & 55.5 & 48.5 & 58.2 & 218.5 & 17.7 & 44.2 & 47.8 & 59.6 & 0.1 & 1.6 & 15.1 & 17.9 \\
(7) kNN Sparsification & 71.2 & 261.6 & 31.1 & 54.9 & 47.8 & 58.0 & 218.0 & 17.1 & 44.2 & 47.8 &63.3 & 1.7 & 5.4 & 22.0 & 20.0 \\
\midrule
\textbf{\ourmodel~(Ours)} &\textbf{72.3} &\textbf{270.1} & \textbf{31.5} & \textbf{55.7} & \textbf{48.6} &\textbf{59.0} & \textbf{220.4} & \textbf{20.3} & \textbf{44.5} & \textbf{48.0} & \textbf{65.8} & 0.8 & \textbf{6.4} & \textbf{23.5} & \textbf{21.1}\\

\bottomrule
\end{tabular}

\noindent
\begin{minipage}[t]{0.43\textwidth}
\vspace{5pt}
\resizebox{\textwidth}{!}{
\begin{tabular}{@{\hskip 0mm}l@{\hskip 0.8mm}|@{\hskip 0.8mm}c@{\hskip 0.8mm}|@{\hskip 0.8mm}c@{\hskip 0.8mm}|@{\hskip 0.8mm}c@{\hskip 0.8mm}|@{\hskip 0.8mm}c@{\hskip 0.8mm}|@{\hskip 0.8mm}c@{\hskip 0.8mm}|@{\hskip 0.8mm}c@{\hskip 0.8mm}}
\toprule
\multicolumn{7}{c}{\textbf{Object-Centric Tasks}} \\
\toprule
 & \multicolumn{3}{c}{\textbf{ScanQA}} & \multicolumn{3}{c}{\textbf{Nr3D}} \\
\cmidrule(lr){2-4} \cmidrule(lr){5-7} 
&  EM1 & M & R & Sim & M & R \\
\midrule
(1) No Vision Tokens &  0& 1.6& 0& 32.0& 10.2& 9.6\\
(2) No Scene Tokens&  15.4& 20.6& 32.1& 44.3& 20.3& 18.7\\
(3) No ROI Tokens &  14.2& 21.5& 34.2& 44.1& 19.0& 18.9\\
(4) Only Vision Tokens  & 10.1 & 14.4 & 23.5 & 44.8 & 19.6 & 17.7\\ 
\midrule
(5) No Depth-Wise CA &  13.9 & 22.4 & 33.9 & 47.9 & \textbf{20.8} & 19.0 \\
(6) No Text-Guidance &  13.6 & 22.2 & 33.5 & 48.2  & \textbf{20.8} &19.1 \\
(7) kNN Sparsification & 14.3 & \textbf{23.9} & \textbf{35.8} & \textbf{48.8} & \textbf{20.8} & 18.7 \\
\midrule
\textbf{\ourmodel~(Ours)} & \textbf{14.4} & 22.9 & 34.8 & 48.2 & \textbf{20.8} & \textbf{19.2} \\
\bottomrule
\end{tabular}}
\end{minipage}
\hfill
\begin{minipage}[t]{0.55\textwidth}
\vspace{15pt}
\caption{Ablation Study of \ourmodel. \textbf{[A] Component Ablation} on removing different token types: (1) Vision tokens absent (text-only input), (2) Task-guided scene tokens absent, (3) Location-guided ROI tokens absent, (4) Prompt tokens absent (vision-only input). \textbf{[B] Task-Guided Sparsifier Architecture Ablation:} (5) The three blocks of cross-attention (CA) are applied only to the final decoder output, not to hidden states, (6) Task prompt-based queries replaced with task-unaware learnable queries, (7) Uniform downsampling replaced with a kNN and attention pooling strategy. All reported metrics are consistent with Tab.~\ref{tab:scene_centric}.}
\label{tab:ablation}
\end{minipage}
\vspace{-5pt}
\end{table*}

\subsection{Ablation Study}
\label{sec:ablation}
To understand the contribution of different components to \ourmodel's performance, we conducted an ablation study. Our analysis reveals that \ourmodel's superior results are primarily due to: (a) the task-guided sparsifier, which leverages global context to provide task-specific scene-level awareness, and (b) the location-guided sparsifier, which offers localized information crucial for object-centric tasks. As shown in Tab. \ref{tab:ablation}, removing either of these modules results in a substantial performance decrease. We further investigated the architecture of the task-guided sparsifier.

\textbf{Task-Guided Sparsifier.} We first examined the impact of task guidance. Replacing text-prompt-based queries with learnable queries caused a substantial performance decrease, especially for scene-centric tasks (Tab. \ref{tab:ablation}),  where the varied nature of prompts necessitates dynamic and task-aware selection of diverse visual cues. Next, we evaluated the benefit of our depth-wise sparsification strategy. Utilizing only the final SceneSplat output, instead of leveraging intermediate decoder features, led to a significant performance drop (Tab. \ref{tab:ablation}) primarily on scene-centric tasks that require the global context provided by earlier decoder layers. Finally, we compared our uniform downsampling strategy to a more advanced language-unaware alternative (attention pooling for early layers, k-NN for the final layer, where spatial information is available). This alternative did not yield improved performance (Tab. \ref{tab:ablation}), confirming the efficiency of our simpler approach without compromising information.

\section{Conclusion}
We introduced \ourmodel, a 3D VLM utilzing  language-aligned Gaussian splats. With \ourmodel, we proposed a paradigm shift in 3D vision-language understanding by moving away from object-centric representations towards a holistic, scene-centric and language-based approach. By directly embedding language features into the spatial structure of 3D scenes, \ourmodel, overcomes the inherent limitations of object detector dependencies, enabling a more natural and comprehensive understanding of complex environments. We also proposed a dual sparsification module that effectively tackles the challenge of dense language-augmented scenes. The task-guided component distills the representation into compact, task-relevant features through task-guided selection on global context. Notably, with \ourmodel, we presented a pioneering 3D VLM operating on Gaussian Splats, leveraging their rich geometric and appearance information for enhanced scene understanding/reasoning tailored to the embodied vision and beyond. Our extensive evaluations across a diverse suite of 3D vision-language tasks demonstrate the clear advantages of our scene-centric approach. \ourmodel~ consistently achieves state-of-the-art performance, significantly outperforming existing methods on scene-centric tasks and also exhibiting strong results on object-centric benchmarks despite being detector-free. Finally, we empirically validated the practical generalization of our method, showing its improved performance on 3D data collected with more readily available equipment.




\section*{ACKNOWLEDGMENT}
This research was partially funded by the Ministry of Education and Science of Bulgaria (support for INSAIT, part of the Bulgarian National Roadmap for Research Infrastructure).



\bibliographystyle{plain}
\bibliography{bibliography}

\begin{thebibliography}{10}

\bibitem{achlioptas2020referit_3d}
Panos Achlioptas, Ahmed Abdelreheem, Fei Xia, Mohamed Elhoseiny, and Leonidas~J. Guibas.
\newblock {ReferIt3D}: Neural listeners for fine-grained 3d object identification in real-world scenes.
\newblock In {\em ECCV}, 2020.

\bibitem{anderson2016spice}
Peter Anderson, Basura Fernando, Mark Johnson, and Stephen Gould.
\newblock Spice: Semantic propositional image caption evaluation.
\newblock In {\em ECCV}, 2016.

\bibitem{scanqa_22_cvpr}
Daichi Azuma, Taiki Miyanishi, Shuhei Kurita, and Motoaki Kawanabe.
\newblock Scanqa: 3d question answering for spatial scene understanding.
\newblock In {\em CVPR}, 2022.

\bibitem{banerjee2005meteor}
Satanjeev Banerjee and Alon Lavie.
\newblock Meteor: An automatic metric for mt evaluation with improved correlation with human judgments.
\newblock In {\em Proceedings of the ACL workshop on intrinsic and extrinsic evaluation measures for machine translation and/or summarization}, 2005.

\bibitem{pizero}
Kevin Black, Noah Brown, Danny Driess, Adnan Esmail, Michael Equi, and et~al.
\newblock $\pi$0: A vision-language-action flow model for general robot control.
\newblock {\em arXiv:2410.24164}, 2024.

\bibitem{NEURIPS2020_1457c0d6}
Tom Brown, Benjamin Mann, Nick Ryder, Melanie Subbiah, Jared~D Kaplan, and et~al.
\newblock Language models are few-shot learners.
\newblock In {\em NeurIPS}, volume~33, 2020.

\bibitem{brown2020language}
Tom Brown, Benjamin Mann, Nick Ryder, Melanie Subbiah, Jared~D Kaplan, et~al.
\newblock Language models are few-shot learners.
\newblock {\em NeurIPS}, 33, 2020.

\bibitem{chen2020scanrefer}
Dave~Zhenyu Chen, Angel~X Chang, and Matthias Nie{\ss}ner.
\newblock Scanrefer: 3d object localization in rgb-d scans using natural language.
\newblock {\em ECCV}, 2020.

\bibitem{chen2024ll3da}
Sijin Chen, Xin Chen, Chi Zhang, Mingsheng Li, Gang Yu, et~al.
\newblock Ll3da: Visual interactive instruction tuning for omni-3d understanding reasoning and planning.
\newblock In {\em CVPR}, 2024.

\bibitem{chen2023end}
Sijin Chen, Hongyuan Zhu, Xin Chen, Yinjie Lei, et~al.
\newblock End-to-end 3d dense captioning with vote2cap-detr.
\newblock In {\em CVPR}, 2023.

\bibitem{chen2024grounded3dllm}
Yilun Chen, Shuai Yang, Haifeng Huang, Tai Wang, Ruiyuan Lyu, et~al.
\newblock Grounded 3d-llm with referent tokens.
\newblock {\em arXiv:2405.10370}, 2024.

\bibitem{chen2021scan2cap}
Zhenyu Chen, Ali Gholami, Matthias Nie{\ss}ner, and Angel~X Chang.
\newblock Scan2cap: Context-aware dense captioning in rgb-d scans.
\newblock In {\em CVPR}, 2021.

\bibitem{chiang2023vicuna}
Wei-Lin Chiang, Zhuohan Li, Ziqing Lin, Ying Sheng, Zhanghao Wu, et~al.
\newblock Vicuna: An open-source chatbot impressing gpt-4 with 90\%* chatgpt quality.
\newblock {\em See https://vicuna. lmsys. org (accessed 14 April 2023)}, 2(3), 2023.

\bibitem{chin2004rouge}
Lin Chin-Yew.
\newblock Rouge: A package for automatic evaluation of summaries.
\newblock In {\em Proceedings of the Workshop on Text Summarization Branches Out, 2004}, 2004.

\bibitem{dai2017scannet}
Angela Dai, Angel~X Chang, Manolis Savva, Maciej Halber, Thomas Funkhouser, and Matthias Nie{\ss}ner.
\newblock Scannet: Richly-annotated 3d reconstructions of indoor scenes.
\newblock In {\em CVPR}, 2017.

\bibitem{Das_2018_CVPR}
Abhishek Das, Samyak Datta, Georgia Gkioxari, Stefan Lee, Devi Parikh, and Dhruv Batra.
\newblock Embodied question answering.
\newblock In {\em CVPR}, 2018.

\bibitem{delitzas2024scenefun3d}
Alexandros Delitzas, Ayca Takmaz, Federico Tombari, Robert Sumner, Marc Pollefeys, et~al.
\newblock {SceneFun3D: Fine-Grained Functionality and Affordance Understanding in 3D Scenes}.
\newblock In {\em CVPR}, 2024.

\bibitem{fu2025scene}
Rao Fu, Jingyu Liu, Xilun Chen, Yixin Nie, and Wenhan Xiong.
\newblock Scene-llm: Extending language model for 3d visual reasoning.
\newblock In {\em IEEE/CVF Winter Conference on Applications of Computer Vision (WACV)}, 2025.

\bibitem{mei2025PerLA}
Mei Guofeng, Lin Wei, Riz Luigi, Wu~Yujiao, Poiesi Fabio, and Wang Yiming.
\newblock Perla: Perceptive 3d language assistant.
\newblock In {\em CVPR}, 2025.

\bibitem{3dllm}
Yining Hong, Haoyu Zhen, Peihao Chen, Shuhong Zheng, Yilun Du, Zhenfang Chen, and Chuang Gan.
\newblock 3d-llm: Injecting the 3d world into large language models.
\newblock {\em NeurIPS}, 2023.

\bibitem{hu2022lora}
Edward~J Hu, Yelong Shen, Phillip Wallis, Zeyuan Allen-Zhu, Yuanzhi Li, et~al.
\newblock Lora: Low-rank adaptation of large language models.
\newblock {\em ICLR}, 1(2), 2022.

\bibitem{huang2024chat}
Haifeng Huang, Yilun Chen, Zehan Wang, Rongjie Huang, Runsen Xu, et~al.
\newblock Chat-scene: Bridging 3d scene and large language models with object identifiers.
\newblock In {\em NeurIPS}, 2024.

\bibitem{huang2023embodied}
Jiangyong Huang, Silong Yong, Xiaojian Ma, Xiongkun Linghu, Puhao Li, et~al.
\newblock An embodied generalist agent in 3d world.
\newblock In {\em ICML}, 2024.

\bibitem{lei2021less}
Jie Lei, Linjie Li, Luowei Zhou, Zhe Gan, Tamara~L Berg, et~al.
\newblock Less is more: Clipbert for video-and-language learning via sparse sampling.
\newblock In {\em CVPR}, 2021.

\bibitem{li2023blip}
Junnan Li, Dongxu Li, Silvio Savarese, and Steven Hoi.
\newblock Blip-2: Bootstrapping language-image pre-training with frozen image encoders and large language models.
\newblock In {\em ICML}, 2023.

\bibitem{li2025m3dbench}
Mingsheng Li, Xin Chen, Chi Zhang, Sijin Chen, Hongyuan Zhu, et~al.
\newblock M3dbench: Towards omni 3d assistant with interleaved multi-modal instructions.
\newblock In {\em ECCV}, 2025.

\bibitem{li2025scenesplat}
Yue Li, Qi~Ma, Runyi Yang, Huapeng Li, Mengjiao Ma, Bin Ren, Nikola Popovic, Nicu Sebe, Ender Konukoglu, Theo Gevers, et~al.
\newblock Scenesplat: Gaussian splatting-based scene understanding with vision-language pretraining.
\newblock {\em arXiv:2503.18052}, 2025.

\bibitem{vis_inst_tuning}
Haotian Liu, Chunyuan Li, Qingyang Wu, and Yong~Jae Lee.
\newblock Visual instruction tuning.
\newblock In A.~Oh, T.~Naumann, A.~Globerson, K.~Saenko, M.~Hardt, and S.~Levine, editors, {\em NeurIPS}, 2023.

\bibitem{lv2024robomp2}
Qi~Lv, Hao Li, Xiang Deng, Rui Shao, et~al.
\newblock Robomp$2$: A robotic multimodal perception-planning framework with mutlimodal large language models.
\newblock In {\em ICML}, 2024.

\bibitem{ma2022sqa3d}
Xiaojian Ma, Silong Yong, Zilong Zheng, Qing Li, Yitao Liang, et~al.
\newblock Sqa3d: Situated question answering in 3d scenes.
\newblock In {\em ICLR}, 2023.

\bibitem{man2024situation}
Yunze Man, Liang-Yan Gui, and Yu-Xiong Wang.
\newblock Situational awareness matters in 3d vision language reasoning.
\newblock In {\em CVPR}, 2024.

\bibitem{papineni2002bleu}
Kishore Papineni, Salim Roukos, Todd Ward, and Wei-Jing Zhu.
\newblock Bleu: a method for automatic evaluation of machine translation.
\newblock In {\em Proceedings of the 40th annual meeting of the Association for Computational Linguistics}, 2002.

\bibitem{Peng2023OpenScene}
Songyou Peng, Kyle Genova, Chiyu~"Max" Jiang, Andrea Tagliasacchi, Marc Pollefeys, and Thomas Funkhouser.
\newblock Openscene: 3d scene understanding with open vocabularies.
\newblock In {\em CVPR}, 2023.

\bibitem{radford2021learning}
Alec Radford, Jong~Wook Kim, Chris Hallacy, Aditya Ramesh, Gabriel Goh, et~al.
\newblock Learning transferable visual models from natural language supervision.
\newblock In {\em ICML}, 2021.

\bibitem{raffel2020exploring}
Colin Raffel, Noam Shazeer, Adam Roberts, Katherine Lee, Sharan Narang, Michael Matena, Yanqi Zhou, Wei Li, and Peter~J Liu.
\newblock Exploring the limits of transfer learning with a unified text-to-text transformer.
\newblock {\em Journal of Machine Learning Research}, 21(140), 2020.

\bibitem{rana2023sayplan}
Krishan Rana, Jesse Haviland, Sourav Garg, Jad Abou-Chakra, et~al.
\newblock Sayplan: Grounding large language models using 3d scene graphs for scalable task planning.
\newblock In {\em CoRL}, 2023.

\bibitem{reimers-gurevych-2019-sentence}
Nils Reimers and Iryna Gurevych.
\newblock Sentence-{BERT}: Sentence embeddings using {S}iamese {BERT}-networks.
\newblock In {\em Empirical Methods in Natural Language Processing}. ACL, 2019.

\bibitem{schonberger2016structure}
Johannes~L Schonberger and Jan-Michael Frahm.
\newblock Structure-from-motion revisited.
\newblock In {\em CVPR}, 2016.

\bibitem{schonberger2016pixelwise}
Johannes~L Sch{\"o}nberger, Enliang Zheng, Jan-Michael Frahm, and Marc Pollefeys.
\newblock Pixelwise view selection for unstructured multi-view stereo.
\newblock In {\em ECCV}, 2016.

\bibitem{shorinwa2024splat}
Ola Shorinwa, Johnathan Tucker, Aliyah Smith, Aiden Swann, et~al.
\newblock Splat-mover: Multi-stage, open-vocabulary robotic manipulation via editable gaussian splatting.
\newblock In {\em CoRL}, 2024.

\bibitem{szot2021habitat}
Andrew Szot, Alex Clegg, Eric Undersander, Erik Wijmans, et~al.
\newblock Habitat 2.0: Training home assistants to rearrange their habitat.
\newblock In {\em NeurIPS}, 2021.

\bibitem{touvron2023llama}
Hugo Touvron, Louis Martin, Kevin Stone, Peter Albert, Amjad Almahairi, et~al.
\newblock Llama 2: Open foundation and fine-tuned chat models.
\newblock {\em arXiv:2307.09288}, 2023.

\bibitem{tschannen2025siglip}
Michael Tschannen, Alexey Gritsenko, Xiao Wang, Muhammad~Ferjad Naeem, Ibrahim Alabdulmohsin, et~al.
\newblock Siglip 2: Multilingual vision-language encoders with improved semantic understanding, localization, and dense features.
\newblock {\em arXiv:2502.14786}, 2025.

\bibitem{vedantam2015cider}
Ramakrishna Vedantam, C~Lawrence~Zitnick, and Devi Parikh.
\newblock Cider: Consensus-based image description evaluation.
\newblock In {\em CVPR}, 2015.

\bibitem{wang2023chat3d}
Zehan Wang, Haifeng Huang, Yang Zhao, Ziang Zhang, and Zhou Zhao.
\newblock Chat-3d: Data-efficiently tuning large language model for universal dialogue of 3d scenes, 2023.

\bibitem{Wijmans_2019_CVPR}
Erik Wijmans, Samyak Datta, Oleksandr Maksymets, Abhishek Das, Georgia Gkioxari, Stefan Lee, Irfan Essa, Devi Parikh, and Dhruv Batra.
\newblock Embodied question answering in photorealistic environments with point cloud perception.
\newblock In {\em CVPR}, 2019.

\bibitem{tidybot}
Jimmy Wu, Rika Antonova, Adam Kan, Marion Lepert, et~al.
\newblock Tidybot: Personalized robot assistance with large language models.
\newblock In {\em IROS}, 2023.

\bibitem{yang20243d}
Dejie Yang, Zhu Xu, Wentao Mo, Qingchao Chen, Siyuan Huang, and Yang Liu.
\newblock 3d vision and language pretraining with large-scale synthetic data.
\newblock {\em arXiv:2407.06084}, 2024.

\bibitem{yeshwanth2023scannet++}
Chandan Yeshwanth, Yueh-Cheng Liu, Matthias Nie{\ss}ner, and Angela Dai.
\newblock Scannet++: A high-fidelity dataset of 3d indoor scenes.
\newblock In {\em ICCV}, 2023.

\bibitem{yu2019deep}
Zhou Yu, Jun Yu, Yuhao Cui, Dacheng Tao, and Qi~Tian.
\newblock Deep modular co-attention networks for visual question answering.
\newblock In {\em CVPR}, 2019.

\bibitem{zhai2023sigmoid}
Xiaohua Zhai, Basil Mustafa, Alexander Kolesnikov, and Lucas Beyer.
\newblock Sigmoid loss for language image pre-training.
\newblock In {\em ICCV}, 2023.

\bibitem{zhang2022opt}
Susan Zhang, Stephen Roller, Naman Goyal, Mikel Artetxe, Moya Chen, et~al.
\newblock Opt: Open pre-trained transformer language models.
\newblock {\em arXiv:2205.01068}, 2022.

\bibitem{zhang2024spartun3d}
Yue Zhang, Zhiyang Xu, Ying Shen, Parisa Kordjamshidi, and Lifu Huang.
\newblock Spartun3d: Situated spatial understanding of 3d world in large language models.
\newblock {\em arXiv:2410.03878}, 2024.

\bibitem{zhi2024lscenellm}
Hongyan Zhi, Peihao Chen, Junyan Li, Shuailei Ma, Xinyu Sun, et~al.
\newblock Lscenellm: Enhancing large 3d scene understanding using adaptive visual preferences.
\newblock {\em arXiv:2412.01292}, 2024.

\bibitem{zhu20233d}
Ziyu Zhu, Xiaojian Ma, Yixin Chen, Zhidong Deng, Siyuan Huang, and Qing Li.
\newblock 3d-vista: Pre-trained transformer for 3d vision and text alignment.
\newblock In {\em CVPR}, 2023.

\bibitem{zhu2025unifying}
Ziyu Zhu, Zhuofan Zhang, Xiaojian Ma, Xuesong Niu, Yixin Chen, et~al.
\newblock Unifying 3d vision-language understanding via promptable queries.
\newblock In {\em ECCV}, 2025.

\end{thebibliography}

\section*{APPENDIX}
This supplementary material provides additional results, implementation details, and data information supporting the main paper. In Sec.~\ref{sup:results}, we present qualitative examples, extended performance comparisons, and ablation studies to further validate the effectiveness of our model. Sec.~\ref{sup:hyperparam} outlines the training and inference configurations used in our experiments. In Sec.~\ref{sup:dataset} we include further dataset information, including a description of the object counting dataset used for the OOD evaluation, along with relevant licensing details on all datasets utilized in this work. Finally, we discuss the limitations of our current approach (Sec.~\ref{sup:limit})and reflect on the broader impact of our work (Sec.~\ref{sup:broader_impact}).

\section{Results and Analysis}
\label{sup:results}
\subsection{Qualitative Results}
\begin{figure*}
    \centering
    \includegraphics[width=\linewidth]{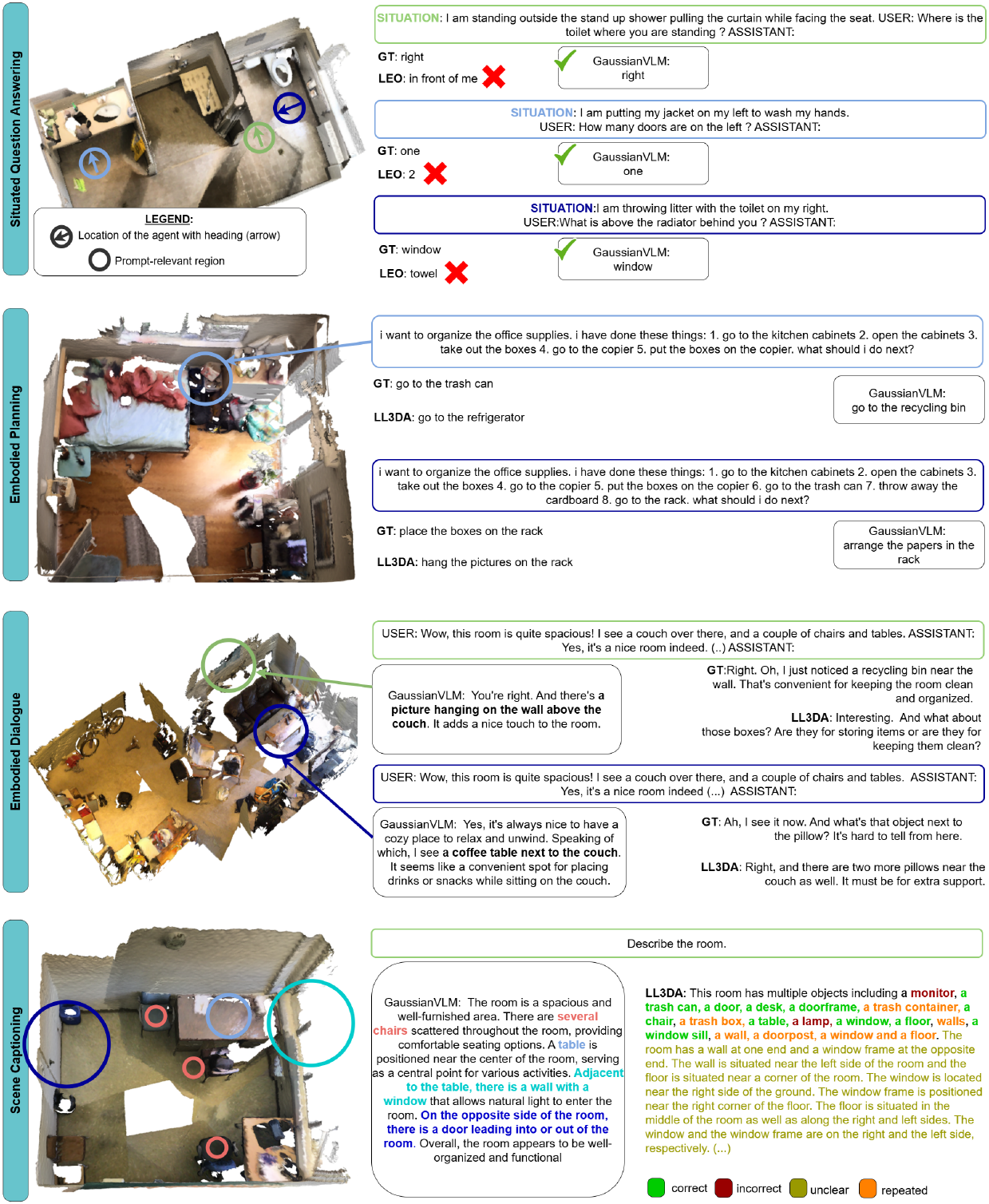}
    \caption{Qualitative results on scene-centric tasks.}
    \label{fig:quali_scene}
\end{figure*}

We present qualitative examples illustrating our model’s performance on both scene-centric and object-centric tasks. As shown in Figure~\ref{fig:quali_scene}, our scene-centric model offers a more comprehensive understanding of the 3D environment. Compared to the baseline, it produces fewer false positives in identified objects and avoids repetitive phrasing in its responses. This indicates not only stronger spatial and contextual grounding, but also higher linguistic fluency and semantic relevance in the generated answers.

\begin{figure*}
    \centering
    \includegraphics[width=\linewidth]{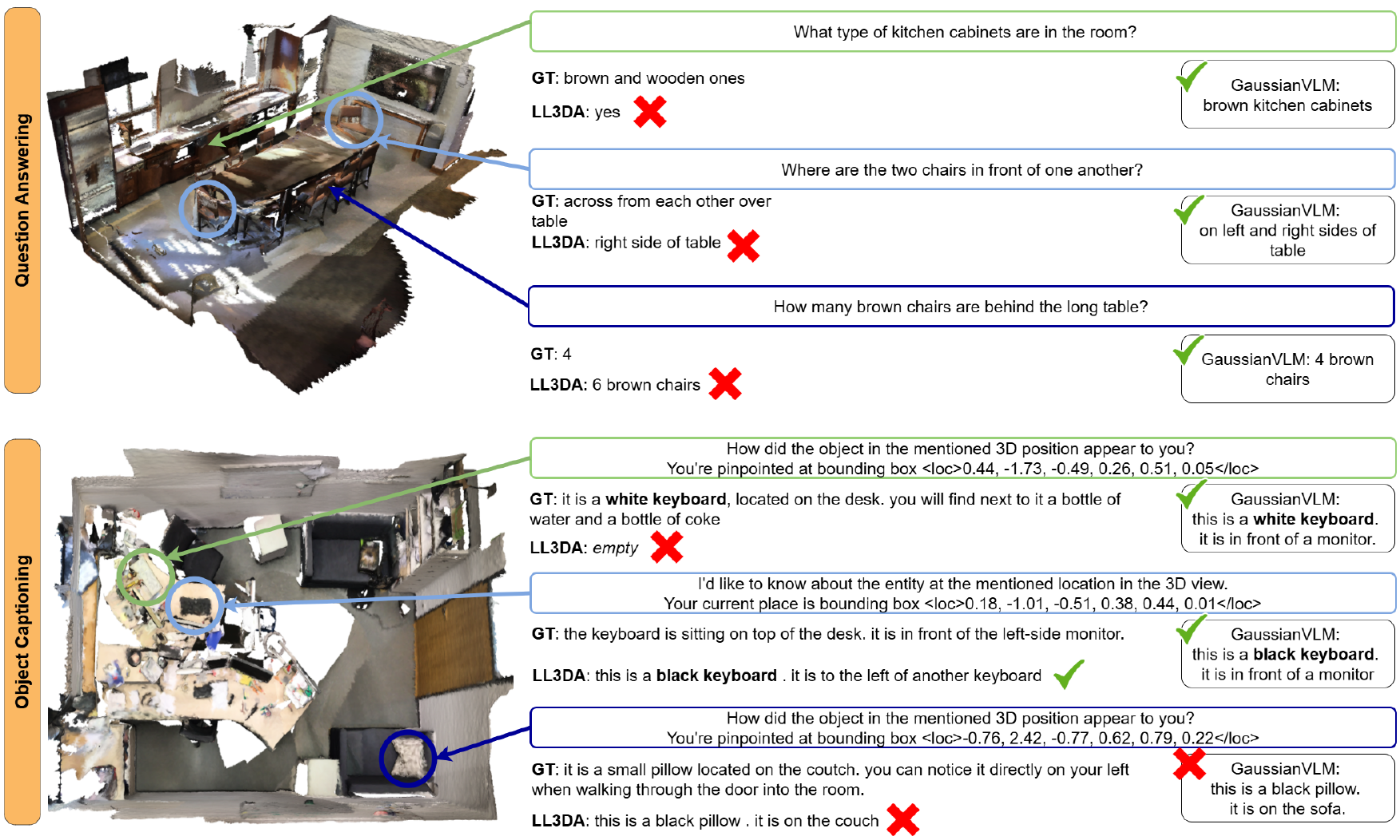}
    \caption{Qualitative results on object-centric tasks.}
    \label{fig:quali_obj}
\end{figure*}

In the object-centric examples shown in Figure~\ref{fig:quali_obj}, the advantages of using Gaussian splats emerge as our model more accurately identifies fine-grained appearance characteristics—such as color, material, or texture. Moreover, in several prompts, the baseline model fails to produce meaningful responses, sometimes returning empty strings or outputs unrelated to the question. In contrast, our model consistently generates relevant and visually grounded answers.

\subsection{Further Results}
We present additional quantitative comparisons to highlight the strengths and limitations of our model, GaussianVLM, across diverse question types. Table~\ref{tab:task_metrics_comparison} shows performance on the ScanQA dataset, disaggregated by question categories. On "How many" questions, which require accurate object counting, LL3DA slightly outperforms GaussianVLM in Exact Match (EM) and ROUGE, though our model achieves a higher METEOR. For "What is" questions, GaussianVLM consistently outperforms LL3DA across all metrics, indicating improved reasoning for open-ended identification tasks. The advantage of our model is more pronounced in "Appearance" questions, where it achieves the highest scores across EM, METEOR, and ROUGE—highlighting the benefit of texture-aware Gaussian Splatting in capturing fine visual details. On the "Where" category, both models struggle in EM, but LL3DA scores higher in the other metrics, possibly due to its design employing object detector, this being more tailored to location-based queries.

In Table \ref{tab:sqa3d_detail}, we compare GaussianVLM against LEO on the SQA3D benchmark across the diverse SQA3D question types. GaussianVLM demonstrates consistent gains in EM and EM-Refined metrics for the majority of question categories, including "What", "Is", "How", and "Others", with improvements up to +5.38 EM-Refined points on "How" questions. These gains suggest our model’s robustness across diverse query types, particularly those requiring spatial-semantic reasoning.

\begin{table*}[ht]
\centering
\begin{tabular}{l|l|c|c|c}
\toprule
\textbf{Task} & \textbf{Method} & \textbf{EM} & \textbf{METEOR} & \textbf{ROUGE} \\
\midrule
How many & LL3DA &  \textbf{0.2723 }&  0.2868 & \textbf{0.5023 }\\
         & GaussianVLM  &  0.2277 &\textbf{0.3039} & 0.4892 \\
\midrule
What is  & LL3DA &  0.1128  & 0.1759 & 0.2647 \\
         & GaussianVLM  &\textbf{0.1142} & \textbf{0.1763} & \textbf{0.2606} \\
\midrule
Appearance    & LL3DA & 0.2862 & 0.2288 & 0.4298 \\
         & GaussianVLM  & \textbf{0.3170}& \textbf{0.2363} & \textbf{0.4565} \\
\midrule
Where    & LL3DA &0.0000 &\textbf{ 0.1911} & \textbf{0.2310} \\
         & GaussianVLM  & 0.0000& 0.1142 & 0.2009 \\
\bottomrule
\end{tabular}
\caption{Performance metrics comparison on different ScanQA categories.}
\label{tab:task_metrics_comparison}
\end{table*}

\begin{table}[ht]
\centering
\begin{tabular}{l|c|c|c}
\toprule
\textbf{Metric} & \textbf{LEO} & \textbf{GaussianVLM} & \textbf{Improvement} \\
\midrule
EM Refined (What)   & 38.45 & \textbf{42.46} & +4.01 \\
EM Refined (Is)     & 63.34 & \textbf{67.94} & +4.60 \\
EM Refined (How)    & 41.72 & \textbf{47.10} & +5.38 \\
EM Refined (Can)    & \textbf{69.23} & 66.27 & -2.96 \\
EM Refined (Which)  & \textbf{48.72} & 46.15 & -2.57 \\
EM Refined (Others) & 49.29 & \textbf{51.41} & +2.12 \\
\midrule
EM (What)           & 33.74 & \textbf{37.49} & +3.75 \\
EM (Is)             & 61.66 & \textbf{65.80} & +4.14 \\
EM (How)            & 41.51 & \textbf{46.88} & +5.37 \\
EM (Can)            & \textbf{69.23} & 66.27 & -2.96 \\
EM (Which)          & \textbf{47.29} & 45.30 & -1.99 \\
EM (Others)         & 44.70 & \textbf{48.94} & +4.24 \\
\bottomrule
\end{tabular}
\caption{Comparison of EM metrics on SQA3D between LEO and GaussianVLM for different question groups. EM-Refined represents an EM adaptation by LEO.}
\label{tab:sqa3d_detail}
\end{table}

\subsection{Ablation Experiments}
\textbf{Minimal ROI Radius.} We conduct an ablation study on the minimal radius used for capturing the ROI, comparing two settings: 15 cm and 30 cm. This analysis is carried out across three benchmarks, two object-centric and one scene-centric. Given that the ROI-based sparsifier is designed to enhance performance on object-centric tasks, we focus primarily on the object-centric tasks, for which we choose ScanRefer and ScanQA. To evaluate its impact on scene-centric performance, we also consider the SQA3D benchmark. The results demonstrate that a smaller ROI of 15 cm significantly benefits object-centric tasks, while only causing a negligible drop in performance on the scene-centric benchmark (Tab.~\ref{tab:ablate_ROI_cm}).

\begin{table}[ht]
\centering
\begin{tabular}{l|c|c}
\toprule
\textbf{Metric} & \textbf{ROI 15} & \textbf{ROI 30} \\
\midrule
\multicolumn{3}{c}{\textbf{ScanRefer}} \\
\midrule
Sentence Similarity & \textbf{0.5914} & 0.5791 \\
\midrule
\multicolumn{3}{c}{\textbf{ScanQA}} \\
\midrule
EM & \textbf{0.1443} & 0.1401 \\
\midrule
\multicolumn{3}{c}{\textbf{SQA3D}} \\
\midrule
EM Overall & 0.4936 & \textbf{0.4942} \\
EM (What) & \textbf{0.3749} & 0.3740 \\
EM (Is) & \textbf{0.6580} & 0.6488 \\
EM (How) & \textbf{0.4688} & 0.4624 \\
EM (Can) & \textbf{0.6627} & 0.6538 \\
EM (Which) & 0.4530 & \textbf{0.5014} \\
EM (Others) & \textbf{0.4894} & 0.4859 \\
\bottomrule
\end{tabular}
\caption{Comparison of GaussianVLM with ROI threshold 15cm and with 30cm.}
\label{tab:ablate_ROI_cm}
\end{table}

\section{Hyperparameter Choice}
\label{sup:hyperparam}
We evaluate our model under two established training protocols from prior state-of-the-art 3D VLMs. We follow the training setup of LL3DA \cite{chen2024ll3da}, and for the embodied reasoning task, SQA3D \cite{ma2022sqa3d}, we adopt the LEO protocol \cite{huang2023embodied}, a benchmarked training strategy for instruction tuning and alignment in 3D vision-language models.

Our architectural hyperparameters are selected based on findings from prior work. We set the number of ROI tokens to 4, aligning with LL3DA’s ablation on location-aware feature encoding using click-based inputs. The number of scene tokens is fixed at 128, based on LSceneLLM \cite{zhi2024lscenellm}, which demonstrates that this token budget provides a good trade-off between performance and efficiency for global scene representations.

The detailed hyperparameter configurations used in our implementation of the LEO training protocol are listed in Tables \ref{tab:leo_hyperparam_1},\ref{tab:leo_hyperparam_2},\ref{tab:leo_hyperparam_3}. Table \ref{tab:leo_hyperparam_1}~ shows the settings for the alignment stage, Table \ref{tab:leo_hyperparam_2}~ for the instruction-tuning stage, and \ref{tab:leo_hyperparam_3}~ for inference. Modifications we made relative to the original LEO setup (e.g., GPU type, output length) are marked with an asterisk. These configurations ensure a consistent and fair comparison while allowing us to scale to high-resolution, language-enriched 3D scenes using Gaussian splats.

\begin{table}[ht]
\centering
\begin{tabular}{l|c}
\toprule
\textbf{Hyperparameter} & \textbf{Value} \\
\midrule
Optimizer & AdamW \\
Weight decay & 0.05 \\
Betas & [0.9, 0.999] \\
Learning rate & 3 $\times$ 10$^{-4}$ \\
Warmup steps & 400 \\
Number of workers & 4 \\
Parallel strategy & DDP \\
Type of GPUs* & NVIDIA A100-80\\
Number of GPUs* & 8 \\
Accumulate gradient batches* & 4 \\
Batch size per GPU & 4  \\
Training precision & bfloat16 \\
Gradient norm & 5.0 \\
Epochs & 5 \\
\bottomrule
\end{tabular}
\caption{Hyperparameters choice of LEO protocol \cite{huang2023embodied} for alignment stage. (*) marks our modifications.}
\label{tab:leo_hyperparam_1}
\end{table}

\begin{table}[ht]
\centering
\begin{tabular}{l|c}
\toprule
\textbf{Hyperparameter} & \textbf{Value} \\
\midrule
Optimizer & AdamW \\
Weight decay & 0.05 \\
Betas & [0.9, 0.999] \\
Learning rate & 3 $\times$ 10$^{-5}$ \\
Warmup steps & 400 \\
Number of workers & 4 \\
Parallel strategy & DDP \\
Type of GPUs* & NVIDIA A100-80 \\
Number of GPUs* & 8 \\
Accumulate gradient batches* & 4 \\
Batch size per GPU & 4 \\
Training precision & bfloat16 \\
Gradient norm & 5.0 \\
Epochs & 10 \\
\bottomrule
\end{tabular}
\caption{Hyperparameters choice of LEO protocol \cite{huang2023embodied} for instruction-tuning stage. (*) marks our modifications.}
\label{tab:leo_hyperparam_2}
\end{table}

\begin{table}[ht]
\centering
\begin{tabular}{l|c}
\toprule
\textbf{Hyperparameters} & \textbf{Value} \\
\midrule
Number of beams & 5 \\
Maximum output length* & 768 \\
Minimum output length & 1 \\
Top p & 0.9 \\
Repetition penalty & 3.0 \\
Length penalty & 1.0 \\
Temperature & 1.0 \\
\bottomrule
\end{tabular}
\caption{Hyperparameters choice of LEO protocol \cite{huang2023embodied} for inference. (*) marks our modifications.}
\label{tab:leo_hyperparam_3}
\end{table}

\section{Datasets Information}
\label{sup:dataset}
In this section, we describe the datasets used for our experiments, spanning both established 3D vision-language datasets as well as a new benchmark introduced in this work for evaluating object counting in real-world 3D reconstructions. We provide details on the construction, purpose, and licensing of the evaluation datasets. In particular, we emphasize the Object Counts Dataset, built on ScanNet++, which allows us to evaluate the generalization of our model to out-of-distribution reconstructions created from RGB data Sec.~\ref{sec:obj_counts}. Table~\ref{tab:dataset-licenses} and Sec.~\ref{sec:license} provide a comprehensive overview of dataset licensing to ensure full transparency and reproducibility.

\subsection{Object Counts Dataset}
\label{sec:obj_counts}
To assess the robustness of \ourmodel~ to deployment scenarios involving diverse data sources, we evaluate its performance on 3D Gaussian splat representations constructed from RGB images captured in real-world environments. Unlike point clouds derived from high-precision laser scanners, RGB-based reconstructions are more accessible and better reflect casual data collection. Our motivation stems from the observation that VLMs trained solely on LiDAR- or scanner-derived point clouds may struggle to generalize to reconstructions without such professional setups. Since our setup comprises only the ScanNet dataset, we leverage the validation split of ScanNet++ -- a high-quality 3D indoor scene dataset collected independently from ScanNet -- for evaluation. We use the Gaussian splat representations of these scenes provided by GaussianWorld, generated from RGB images, and compare against the COLMAP-derived point clouds available for ScanNet++. To the best of our knowledge, no object captioning or question answering benchmarks exist for this dataset. To address this, we construct a new benchmark focused on object counting. Using ScanNet++ segmentation annotations, we automatically extract instance counts and generate 1,000 question-answer pairs of the form "How many \textless label\textgreater{} are in the scene?", with 10 synonym question variants and 5 possible answer rephrasings per instance (e.g., "3", "3 chairs", "I can count 3", etc.). Labels corresponding to non-countable "stuff" categories (e.g., wall, floor, windowsill) and artifacts (e.g., "SPLIT", "REMOVE") are excluded.  

\begin{figure*}
    \centering
    \includegraphics[width=0.9\linewidth]{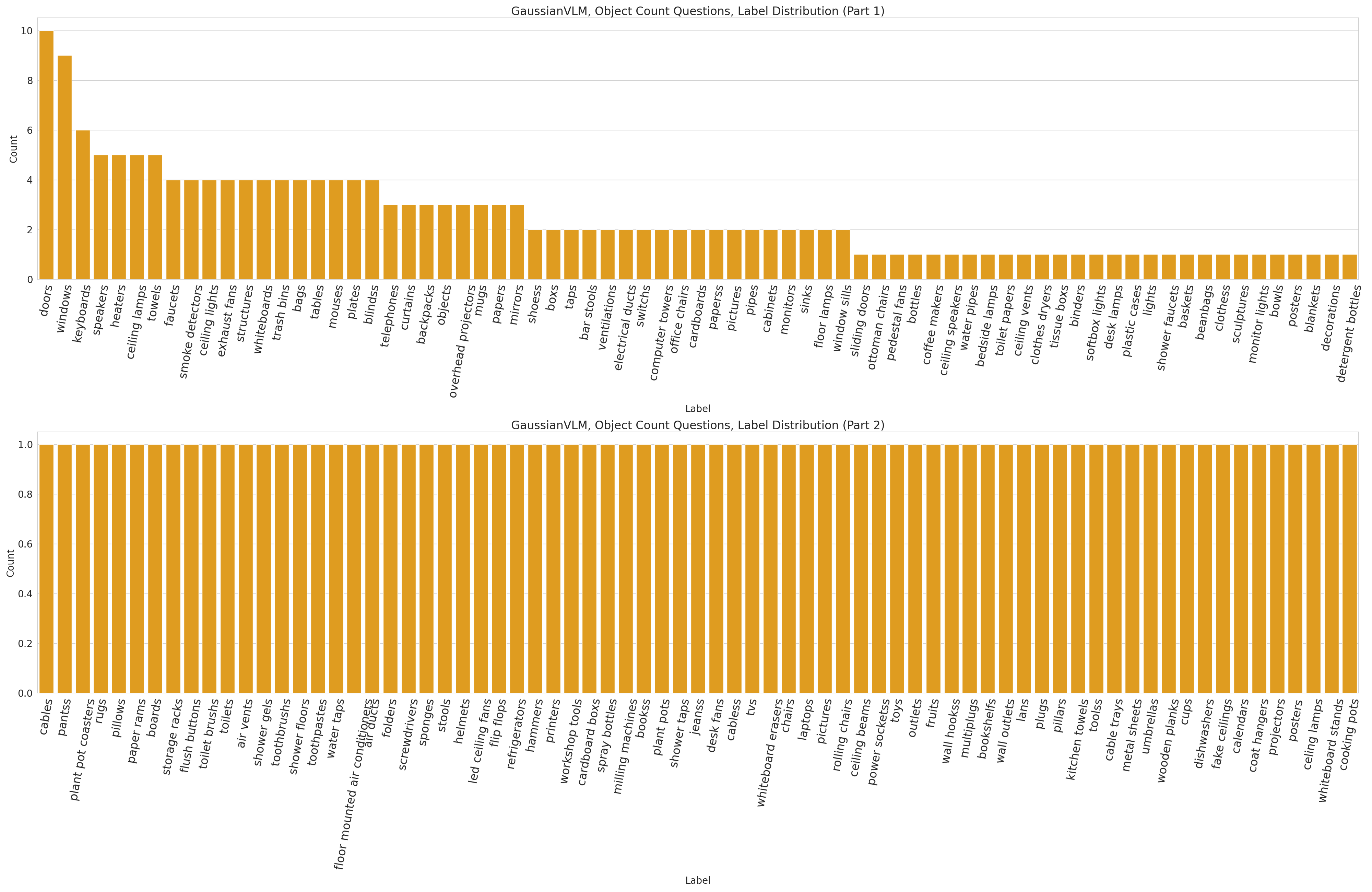}
    \caption{Distribution of the questions on object counts, answered \textbf{correctly} by \textbf{\ourmodel}. The distribution is according to \textbf{object class labels}. Overall, 254 questions answered correctly.}
    \label{fig:dist_labels_ours}
\end{figure*}

\begin{figure*}
    \centering
    \includegraphics[width=0.9\linewidth]{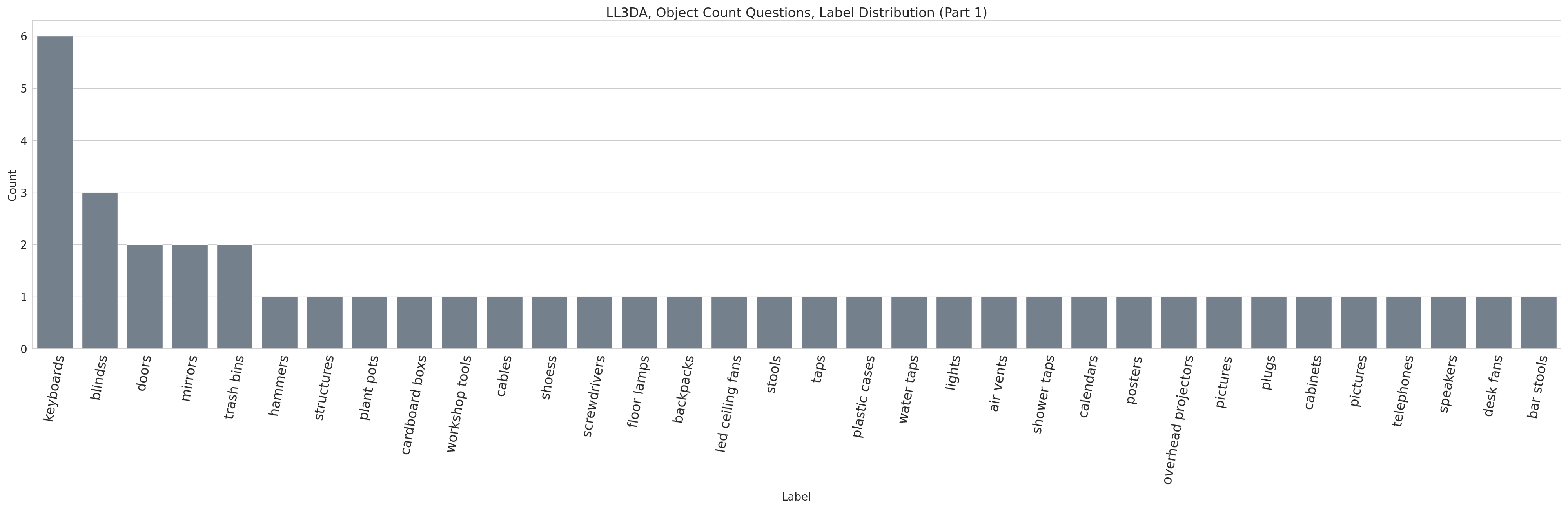}
    \caption{Distribution of the questions on object counts, answered \textbf{correctly} by \textbf{LL3DA}. The distribution is according to \textbf{object class labels}. Overall, 44 questions answered correctly.}
    \label{fig:dist_labels_ll3da}
\end{figure*}

\begin{figure*}
    \centering
    \includegraphics[width=0.8\linewidth]{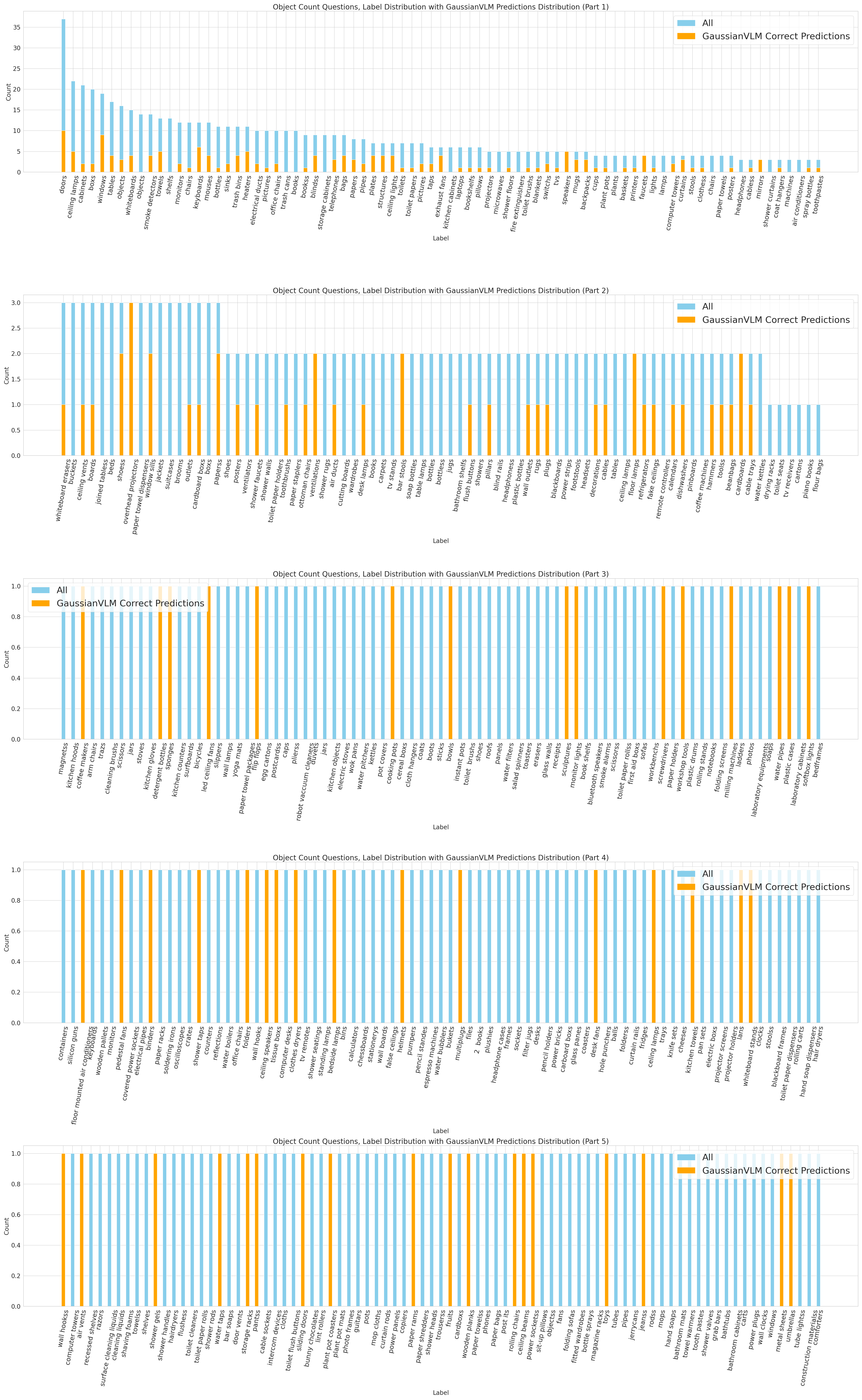}
    \caption{Distribution of object count questions (correcly answered by \ourmodel, vs all questions) according to \textbf{object class labels}.}
    \label{fig:ours_overlayed_class}
\end{figure*}

\begin{figure*}
    \centering
    \includegraphics[width=0.8\linewidth]{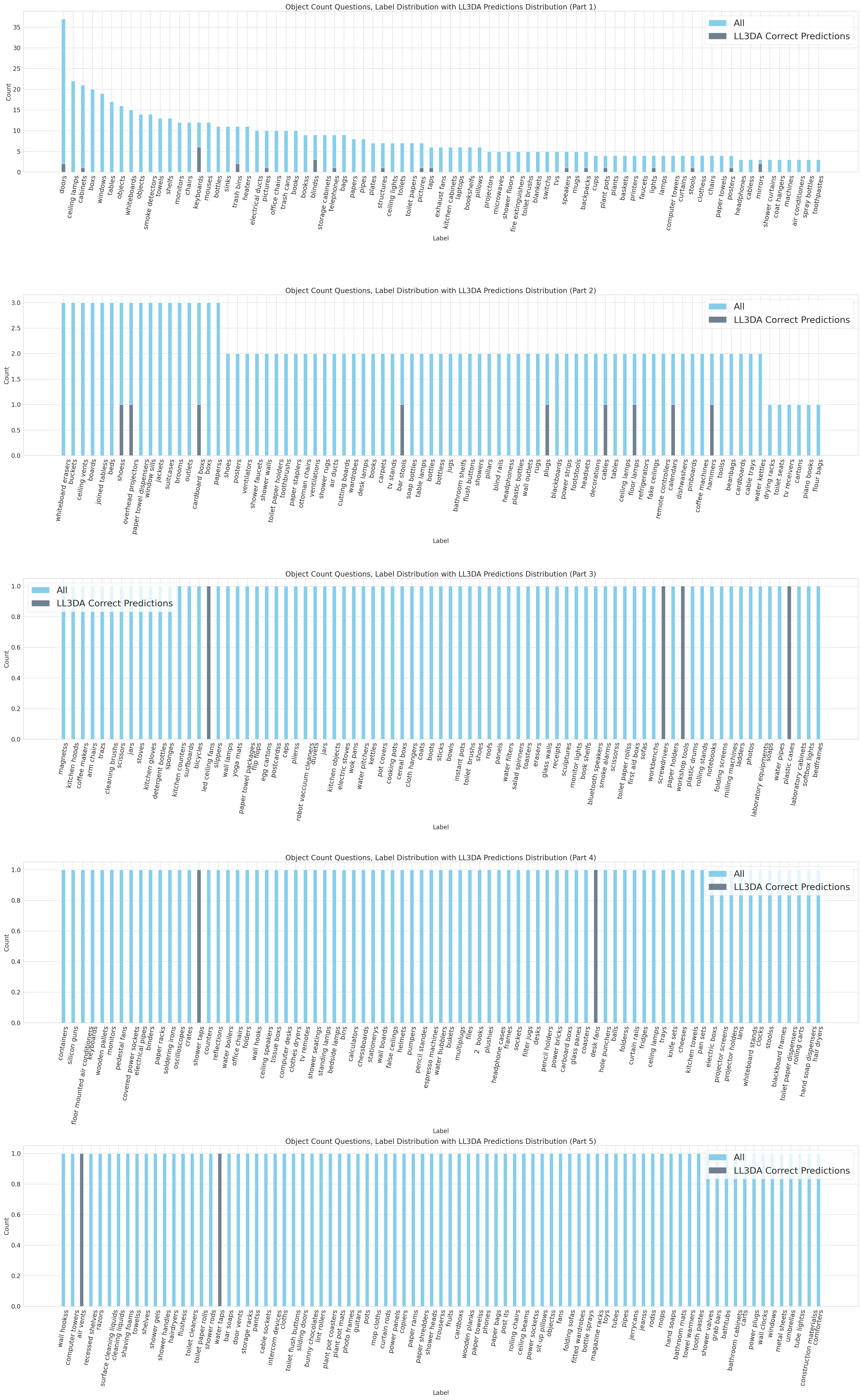}
    \caption{Distribution of object count questions (correcly answered by LL3DA, vs all questions) according to \textbf{object class labels}.}
    \label{fig:ll3da_overlayed_class}
\end{figure*}

Figures~\ref{fig:ours_overlayed_class} and ~\ref{fig:ll3da_overlayed_class} visualize the distribution of object count questions across object class labels for all 1,000 questions, overlaid with those correctly answered by  \ourmodel~ and LL3DA, respectively. These show that \ourmodel~ answers correctly across a wider range of object types. Complementary Figures~\ref{fig:dist_labels_ours} and ~\ref{fig:dist_labels_ll3da} break this down for  \ourmodel~ (254 correct answers) and LL3DA (44 correct answers), highlighting the per-class accuracy gap.

\begin{figure*}
    \centering
    \includegraphics[width=0.9\linewidth]{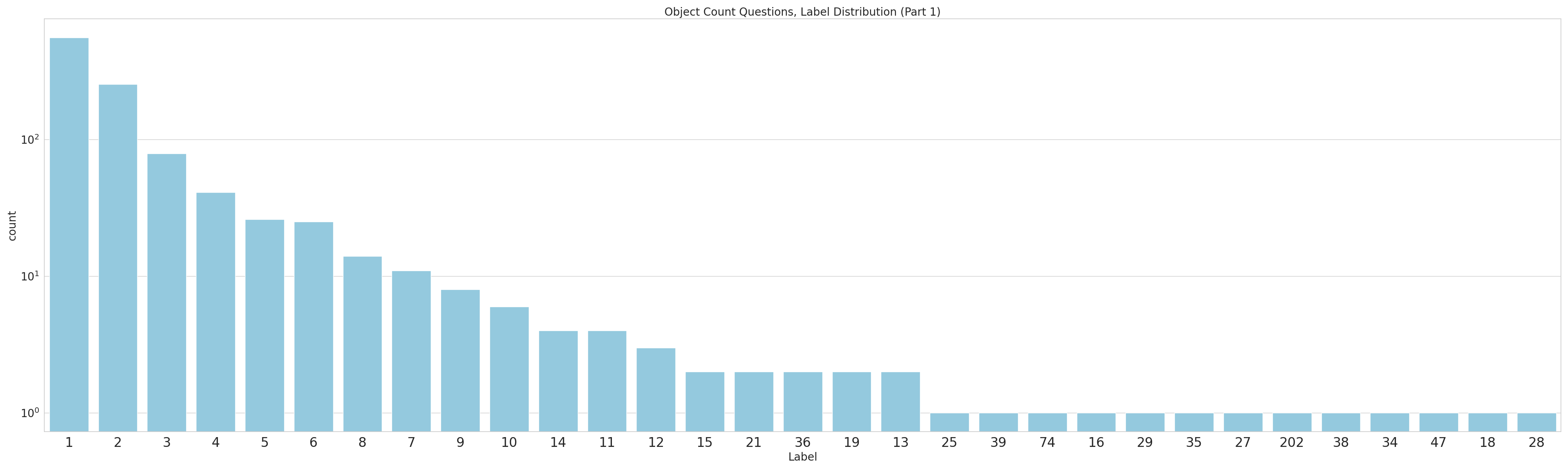}
    \vspace{-10pt}
    \caption{Distribution of object count questions according to \textbf{object count labels}.}
    \label{fig:all_counts}
\end{figure*}

\begin{figure*}
    \centering
    \includegraphics[width=0.9\linewidth]{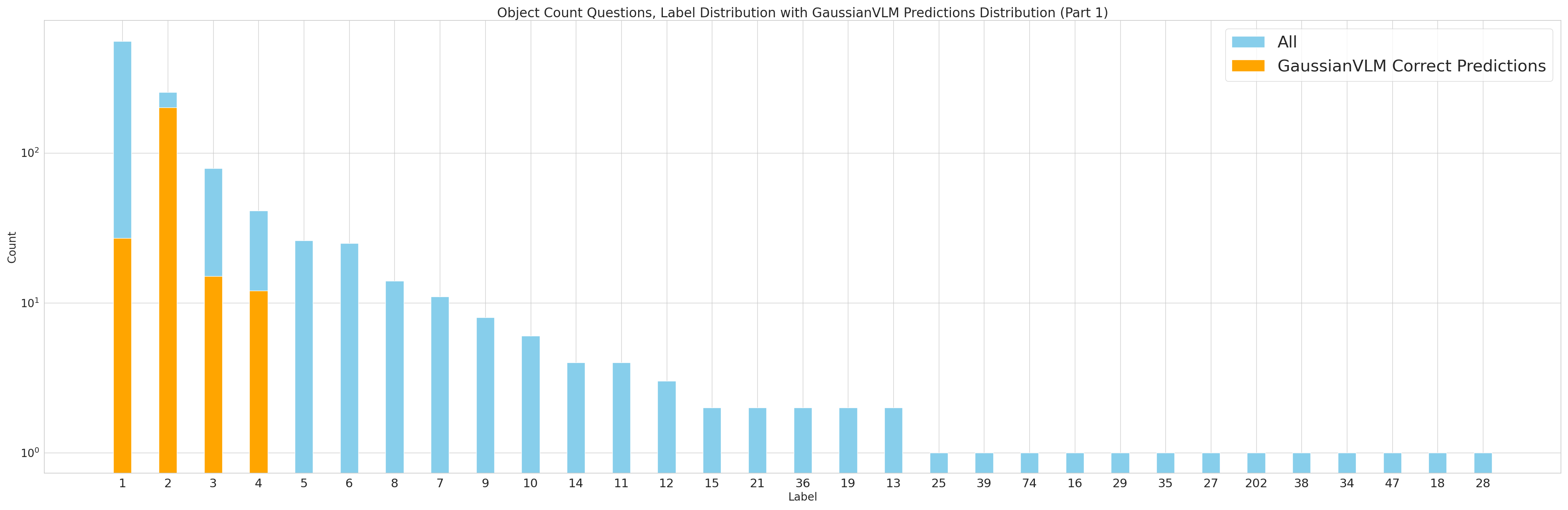}
    \vspace{-10pt}
    \caption{Distribution of object count questions (correcly answered by GaussianVLM, vs all questions) according to \textbf{object count labels}. Overall, 254 questions answered correctly. Logarithmic scaling for the distribution.}
    \label{fig:ours_overlayed}
\end{figure*}

\begin{figure*}
    \centering
    \includegraphics[width=0.9\linewidth]{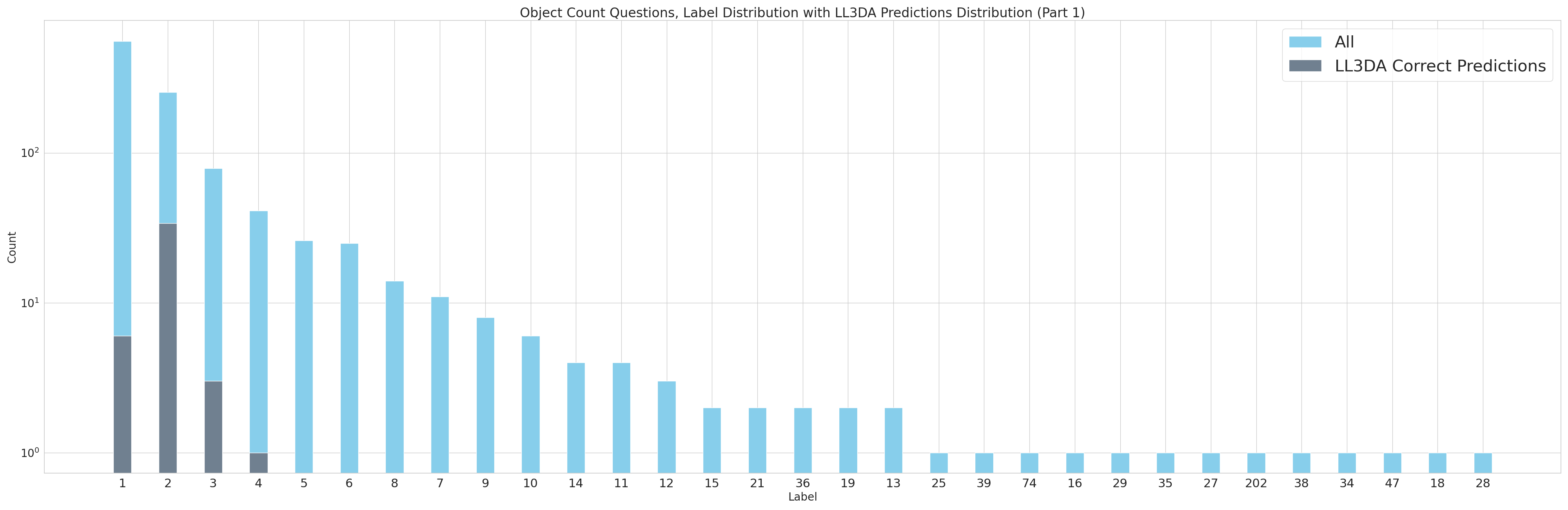}
    \vspace{-10pt}
    \caption{Distribution of object count questions (correcly answered by LL3DA, vs all questions) according to \textbf{object count labels}. Overall, 44 questions answered correctly. Logarithmic scaling for the distribution.}
    \label{fig:ll3da_overlayed}
\end{figure*}

Separately, Figures~\ref{fig:ours_overlayed} and~\ref{fig:ll3da_overlayed} show the distribution of questions by object count values (e.g., “1 chair”, “5 doors”), again overlaid with correctly answered instances. These plots demonstrate that \ourmodel~ generalizes well across a broader range of object counts, including mid-to-high cardinalities, while LL3DA struggles with higher counts. Figure~\ref{fig:all_counts} shows the global distribution of object counts in the benchmark, confirming that the dataset includes a wide and balanced spectrum of count values.

We evaluate on both LL3DA and GaussianVLM; note that novel data evaluation is limited to LL3DA due to LEO’s lack of an integrated object detector enabling evaluation on further datasets. We evaluate using standard QA metrics -- exact match, ROUGE, METEOR, CIDEr, BLEU -- and a custom accuracy metric. Accuracy accounts for rephrasings and approximate number matching by extracting numeric tokens from predictions and ground truths (including both digit and word forms) and comparing them after normalization, regardless of the sentence context.

All data will be made publicly available.


\begin{figure*}
    \centering
    \includegraphics[width=0.9\linewidth]{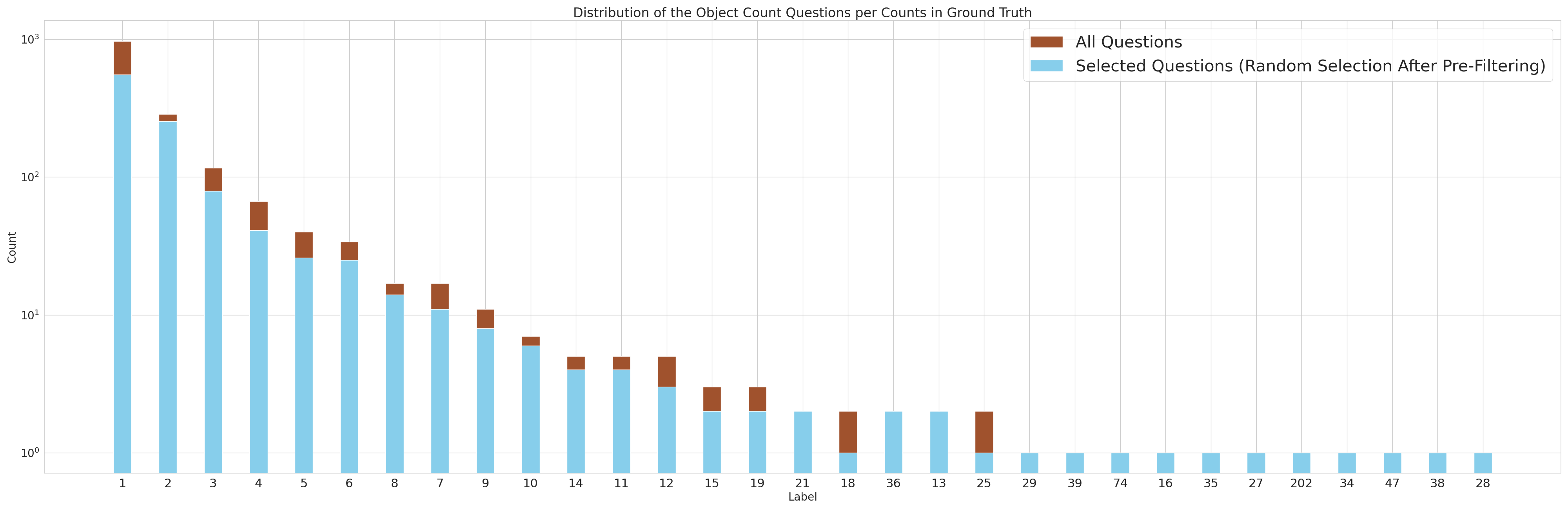}
    \vspace{-10pt}
    \caption{Distribution of object count questions based on ground truth object counts labels (log scale). We show the initial distribution upon generating the questions (red) and the distribution of the questions used in our evaluations (blue).}
    \label{fig:all_selected_counts}
\end{figure*}

\begin{figure*}
    \centering
    \includegraphics[width=0.8\linewidth]{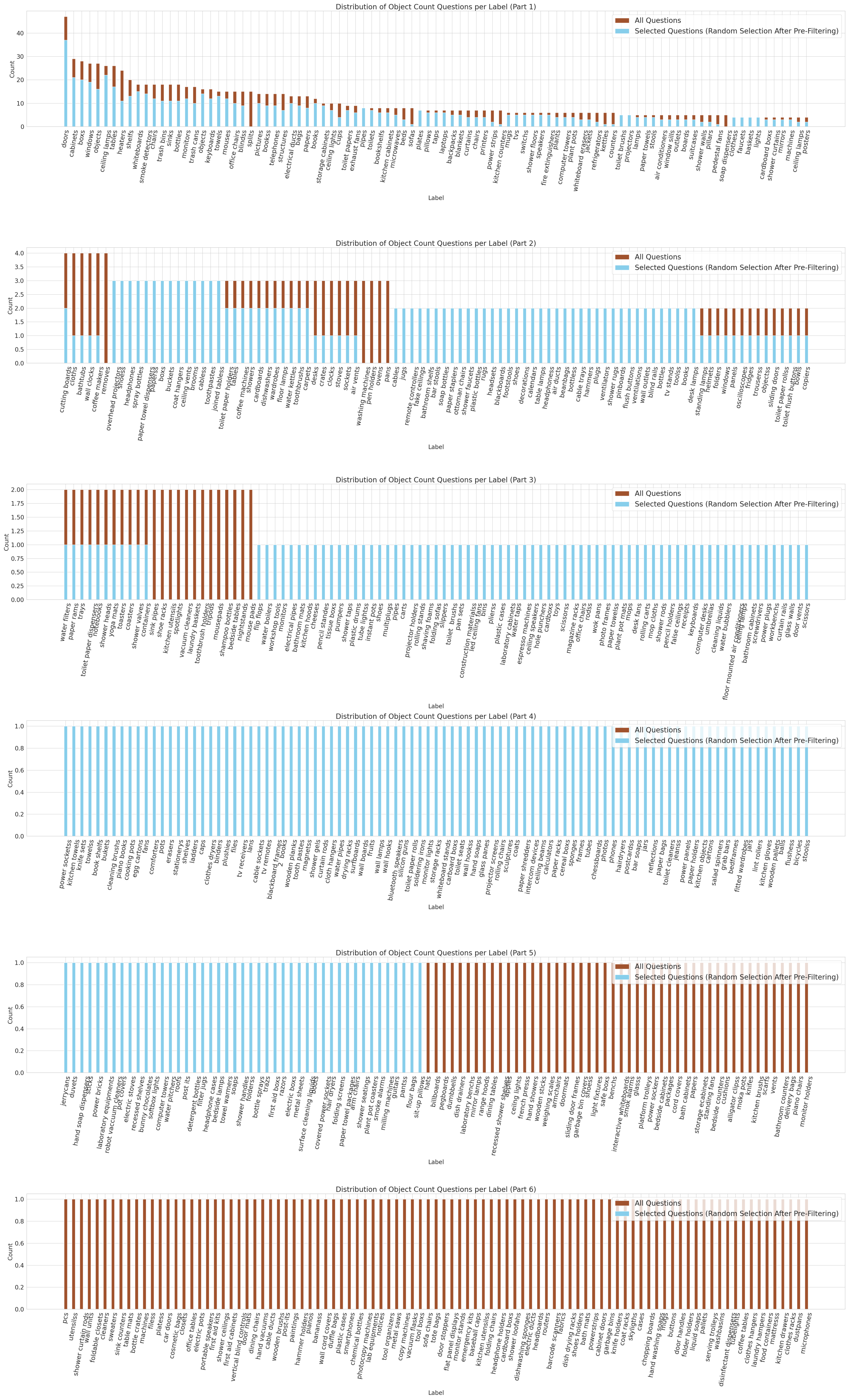}
    \caption{Distribution of object count questions based on object type label. We show the initial distribution upon generating the questions (red) and the distribution of the questions used in our evaluations (blue).}
    \label{fig:all_selected_labels}
\end{figure*}

\subsection{Dataset Licenses}
\label{sec:license}
\begin{table}[h!]
\centering
\small
\begin{tabular}{p{2.9cm}|p{4.5cm}}
\toprule
\textbf{Dataset} & \textbf{License / Terms of Use} \\
\midrule
ScanNet \cite{dai2017scannet} & ScanNet Terms of Use\footnotemark[1] \\
ScanNet++ \cite{yeshwanth2023scannet++} & ScanNet++ Terms of Use\footnotemark[2] \\
ScanRefer \cite{chen2020scanrefer} & Creative Commons BY-NC-SA 3.0 \\
ScanQA \cite{scanqa_22_cvpr} & Apache 2.0 \\
SQA3D \cite{ma2022sqa3d}& Apache 2.0 \\
ReferIt3D \cite{achlioptas2020referit_3d}& MIT \\
3D-LLM \cite{3dllm}& MIT \\
GaussianWorld (SceneSplat-7K) \cite{li2025scenesplat}& Same as ScanNet, 
ScanNet++ \\
\bottomrule
\end{tabular}
\caption{Licenses and terms of use for datasets employed in our study.}
\label{tab:dataset-licenses}
\end{table}

We summarize the licenses and terms of use for all datasets used in this work in Table~\ref{tab:dataset-licenses}. All datasets are publicly released, and we adhere strictly to the respective terms. Notably, ScanNet \cite{dai2017scannet} and ScanNet++ \cite{yeshwanth2023scannet++} are governed by their own custom terms of use, while other datasets adopt standard open-source licenses. GaussianWorld (SceneSplat-7K) inherits licensing from the datasets it reprocesses -- in our case, ScanNet and ScanNet++ -- and therefore follows the same terms.

\footnotetext[1]{Available at \url{https://kaldir.vc.in.tum.de/scannet/ScanNet_TOS.pdf} (last accessed: 19/05/2025).}
\footnotetext[2]{Available at \url{https://kaldir.vc.in.tum.de/scannetpp/static/scannetpp-terms-of-use.pdf} (last accessed: 19/05/2025).}

\section{Limitations}
\label{sup:limit}
While \ourmodel~ demonstrates strong generalization and performance across a variety of 3D vision-language tasks, several limitations remain.

First, although our model maintains computational parity with other recent 3D VLMs -- measured in total training hours under comparable training protocols -- the broader class of vision-language models for 3D reasoning remains computationally intensive. As a result, real-time or resource-constrained inference may still pose practical challenges. While our multi-phase, multi-branch sparsification strategy is specifically designed to reduce computational bottlenecks, the underlying 3D backbone architecture, though SOTA, remains heavy.

Second, training these VLMs is also resource-intensive, requiring a dedicated A100-80 node (8 GPUs) for 24 hours.

Third, our evaluation focuses on static 3D scenes. Dynamic or multi-agent environments -- common in robotics and AR/VR -- are not addressed. Extending the model to handle time-varying inputs and temporal reasoning is a potential direction for future work.

Fourth, while Gaussian splatting enables realistic reconstructions from RGB, the quality and completeness of reconstructions can vary significantly depending on the capture process. Our experiments on ScanNet++ assume clean reconstructions; model performance may degrade on lower-quality or outdoor scenes.

Finally, our object counting benchmark on ScanNet++ covers only one type of task in the out-of-distribution (OOD) evaluation setting. A broader set of benchmarks across diverse OOD conditions is necessary to fully assess the generalization of 3D VLMs to unconstrained environments.

\section{Broader Impact}
\label{sup:broader_impact}
Our work aims to expand the capabilities of vision-language models (VLMs) for holistic 3D scene understanding, moving beyond object-centric paradigms that rely heavily on predefined taxonomies and bounding-box supervision. By leveraging scene-centric representations and operating directly on expressive 3D inputs such as Gaussian splats, our approach offers potential benefits for real-world applications that require open-ended, spatially grounded reasoning, such as robotics, assistive technologies, and AR/VR systems.

However, our approach also comes with potential environmental implications. Although our model is designed with efficiency in mind -- via sparsification and modularization -- training large-scale VLMs, including our own, still requires significant computational resources and GPU hours. This high energy consumption contributes to environmental concerns, showing the need for future research on  methods that reduce training footprints.



\end{document}